\documentclass[runningheads]{llncs}
\usepackage{etex}
\usepackage{url}
\usepackage[numbers]{natbib}
\renewcommand\bibname{References}
\usepackage{cvpr-abbriv}
\usepackage{epsfig}
\usepackage{graphicx}
\usepackage{amsmath}
\usepackage{amssymb}
\usepackage{array}
\usepackage{tabularx}
\usepackage{url}
\usepackage{graphicx}
\usepackage{amsmath,amssymb} 
\makeatletter
\let\proof\@undefined                        
\let\endproof\@undefined                  
\makeatother

\usepackage{amsthm}
\usepackage{amsthm}
\usepackage{amsxtra}
\usepackage{stmaryrd}
\usepackage{comment}
\usepackage[ruled,vlined,linesnumbered,procnumbered,longend]{algorithm2e}

\SetCommentSty{mycommfont}
\usepackage{multirow,rotating}
\usepackage{array}
\usepackage{afterpage}
\usepackage{dblfloatfix}
\usepackage{setspace}
\usepackage{floatrow}
\usepackage{booktabs}
\usepackage{color, colortbl}
\usepackage{tabularx,calc}
\usepackage{tikz}                 
\usetikzlibrary{patterns,fit,external}
\usepackage{pgfplots}             
\usepackage{overpic}
\usepackage{bbm}
\usepackage[absolute]{textpos}
\usepackage{atbeginend}
\usepackage{units}
\usepackage{xifthen}
\usepackage{chngcntr}
\usepackage{listings}

\usepackage{times}
\usepackage{tikzscale}
\usepackage{marginnote}
\usepackage[textwidth=2cm,figwidth=\columnwidth,colorinlistoftodos]{todonotes}
\makeatletter
\renewcommand{\todo}[2][]{\tikzexternaldisable\@todo[#1]{#2}\tikzexternalenable}
\makeatother

\newcounter{mycomment} 

\usepackage{xifthen}

\usepackage[draft=false, pagebackref=false,breaklinks=true,colorlinks,bookmarks=false,pdftex]{hyperref}
\hypersetup{hidelinks,colorlinks=false,linkbordercolor={1 1 1}}
\DeclareMathAlphabet{\mathcalligra}{T1}{calligra}{m}{n}
\DeclareMathAlphabet{\mathantt}{OT1}{antt}{li}{it}
\DeclareMathAlphabet{\mathpzc}{OT1}{pzc}{m}{it}

\renewcommand{\mid}{\,|\,}

\DeclareMathOperator{\E}{\mathop{\mathbb{E}}}

\newcommand{\N}{\mathcal{N}}

\def\T{^{\mathsf T}}

\newcounter{myRomanCounter}

\newcommand{\x}{{x}}

\newcommand{\w}{{w}}

\def\c{\textsc{c}}

\newcommand{\gray}{\color[rgb]{0.5,0.5,0.5}}
\newcommand{\red}{\color[rgb]{1,0,0}}

\renewcommand*{\paragraph}[1]{\par\noindent{\normalsize\bf #1}\,\xspace}

\def\ij{i,j}

\def\epsilon{\varepsilon}

\newcommand{\AND}{\mbox{ \rm and }}

\makeatletter
\let\parSym\S
\def\S{\mathcal{S}}
\makeatother

\newcommand{\larr}[1]{{\begin{subarray}{l}\setlength{\arraycolsep}{0.2em}#1\end{subarray}}}



\def\cf{\emph{cf}\onedot} 

\newcommand{\revisit}[1][]{%
\ifthenelse{\equal{#1}{}}{
\ensuremath{\red \triangle}\xspace}{%
{\ensuremath{\red \rhd}\xspace}%
{\gray #1}%
{\ensuremath{\red \lhd}\xspace}%
}%
}

\def\anchor [#1]#2{%
\phantomsection{}#1\label{#2}%
\def\arga{#2}%
\global\expandafter\def\csname#2\endcsname{%
\hyperref[#2]{#1}\xspace%
}%
}%

\def\codefunction [#1]#2{%
\phantomsection{}\label{#2}{\ttfamily #1\xspace}%
\def\arga{#2}%
\global\expandafter\def\csname#2\endcsname{%
\hyperref[#2]{\ttfamily #1}\xspace%
}%
}

\usepackage{array}
\newcolumntype{L}[1]{>{\raggedright\let\newline\\\arraybackslash\hspace{0pt}}m{#1}}
\newcolumntype{C}[1]{>{\centering\let\newline\\\arraybackslash\hspace{0pt}}m{#1}}
\newcolumntype{R}[1]{>{\raggedleft\let\newline\\\arraybackslash\hspace{0pt}}m{#1}}

\SetKwFor{forever}{while true}{}{end while}

\def\epsilon{\varepsilon}

\def\={\,{=}\,}

\newlength{\myskip}
\SetAlgoSkip{skipalgomyskip}
\SetAlgoInsideSkip{}

\makeatletter
\let\corollary\@undefined
\let\c@corollary\@undefined
\let\endcorollary\@undefined
\let\definition\@undefined
\let\c@definition\@undefined
\let\enddefinition\@undefined
\let\proof\@undefined
\let\endproof\@undefined
\let\theorem\@undefined
\let\c@theorem\@undefined
\let\endtheorem\@undefined
\let\lemma\@undefined
\let\c@lemma\@undefined
\let\endlemma\@undefined
\let\example\@undefined
\let\c@example\@undefined
\let\endexample\@undefined
\let\remark\@undefined
\let\c@remark\@undefined
\let\endremark\@undefined
\let\proposition\@undefined
\let\c@proposition\@undefined
\let\endproposition\@undefined
\let\property\@undefined
\let\endproperty\@undefined
\makeatother

\usepackage{thmtools}

\newtheoremstyle{tightItalic}
  {0.5\myskip}
  {0\myskip}
  {}
  {}
  {\itshape}
  {.}
  { }
  {}

\newtheoremstyle{tightBf}
  {0.5\myskip}
  {0.5\myskip}
  {}
  {}
  {\bf}
  {.}
  {.5em}
  {}

\theoremstyle{definition}
\theoremstyle{tightBf}

\declaretheorem[style=tightBf,name=Theorem]{theorem}
\declaretheorem[style=tightBf,name=Lemma]{lemma}
\declaretheorem[style=tightBf,name=Definition]{definition}
\declaretheorem[style=tightBf,name=Corollary]{corollary}
\declaretheorem[style=tightBf,name=Proposition]{proposition}

\declaretheorem[style=tightBf,name=Example]{example}

\declaretheorem[style=tightItalic,name=Remark]{remark}

\theoremstyle{tightItalic}
\newtheorem*{proof}{Proof}
\BeforeEnd{proof}{\qed}


\usepackage{thmtools, thm-restate}
\usepackage[capitalise,nameinlink]{cleveref}
\crefformat{equation}{(#2#1#3)}
\crefname{section}{\parSym}{\parSym\parSym}
\Crefname{section}{\parSym}{\parSym\parSym}
\crefname{appendix}{\parSym}{\parSym\parSym}
\crefname{observation}{Observation}{Observations}

\usepackage[bottom]{footmisc}

\makeatletter
\renewcommand\tableofcontents{%
    \@starttoc{toc}%
}
\makeatother

\newcommand{\norma}[1]{\lVert#1\rVert}
\begin{document}
\title{Stochastic Normalizations as Bayesian Learning}
\titlerunning{Stochastic Normalizations as Bayesian Learning}
\author{Alexander Shekhovtsov \and Boris Flach}
\authorrunning{A. Shekhovtsov and B. Flach}
\institute{Czech Technical University in Prague, Dept.~of Cybernetics}
\maketitle

\begin{abstract}
In this work we investigate the reasons why Batch Normalization (BN) improves the generalization performance of deep networks. We argue that one major reason, distinguishing it from data-independent normalization methods, is randomness of batch statistics. This randomness appears in the parameters rather than in activations and admits an interpretation as a practical Bayesian learning. We apply this idea to other (deterministic) normalization techniques that are oblivious to the batch size. We show that their generalization performance can be improved significantly by Bayesian learning of the same form. We obtain test performance comparable to BN and, at the same time, better validation losses suitable for subsequent output uncertainty estimation through approximate Bayesian posterior.
\end{abstract}

\section{Introduction}
Recent advances in hardware and deep NNs make it possible to use large capacity networks, so that the training accuracy becomes close to 100\% even for rather difficult tasks. At the same time, however, we would like to ensure small generalization gaps, \ie a high validation accuracy and a reliable confidence prediction. For this reason, regularization methods become very important.

\begin{figure}[!t]
\centering
\setlength{\tabcolsep}{2pt}
\resizebox{\linewidth}{!}{
\begin{tabular}{cc}
\ \ \ \ \footnotesize{Training Loss} & \ \ \ \ \footnotesize Validation Accuracy \\
\includegraphics[width=0.5\linewidth]{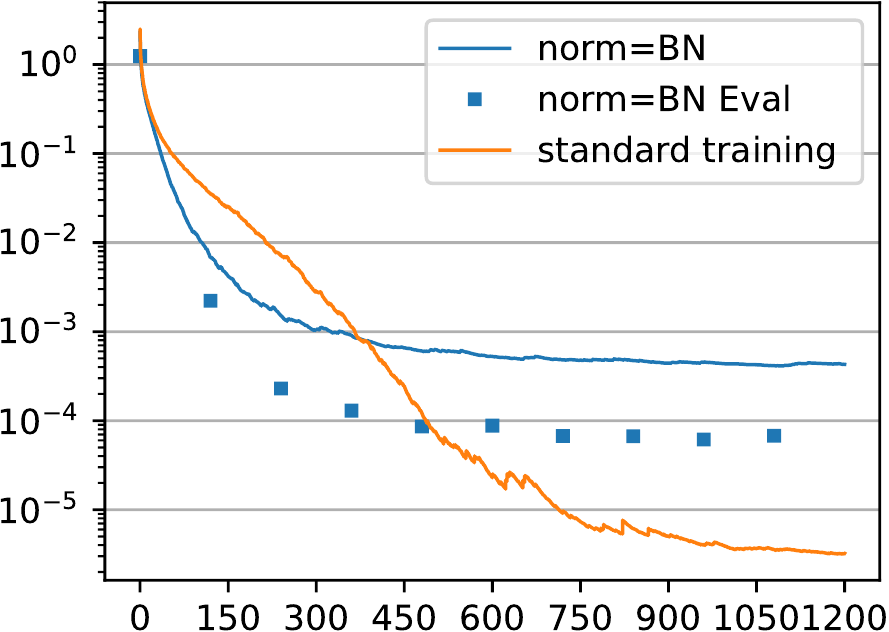}&%
\includegraphics[width=0.5\linewidth]{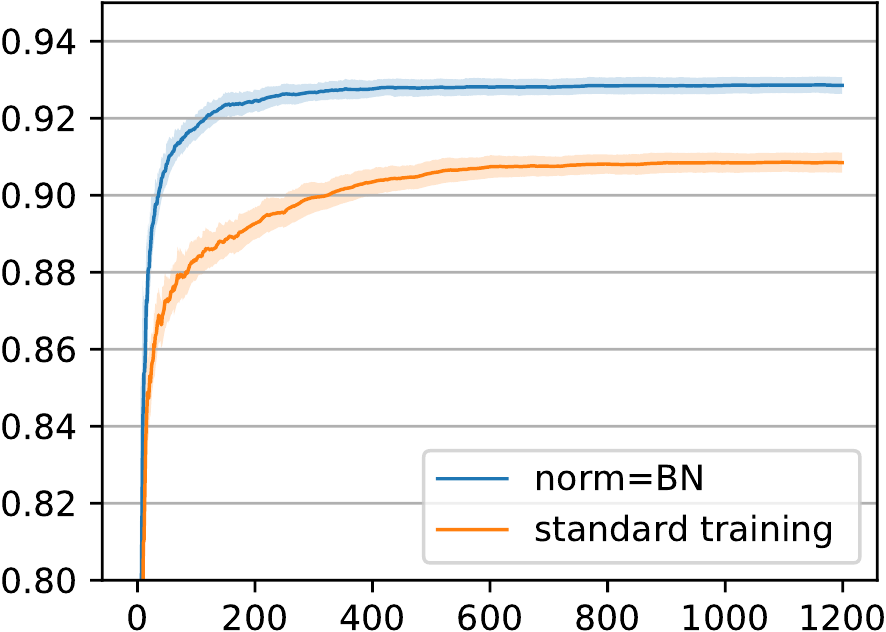}\\
\end{tabular}
}
\caption{The regularization effect of BN during the training epochs (x-axis).
{\em Left:} Batch Normalization, designed for better and faster minimization of the training loss, converges to a small but non-zero value.
Standard training reaches numerical accuracy. When BN is switched to the evaluation mode (BN Eval) the loss on the {\em same} training data is by an order of magnitude smaller, \ie, it shows generalization when switching the mode. {\em Right:} BN clearly achieves a higher validation accuracy. The learning rate is chosen by numerically optimizing the training loss in 5 epochs for each method. Please refer to~\cref{sec:experiment} for details regarding the experimental setup.
\label{fig:BN-regularizes}}
\end{figure}

As the base model for this study we have chosen the All-CNN network of~\cite{Springenberg-14}, a network with eight convolutional layers, and train it on the CIFAR-10 dataset. Recent work~\cite{Gast18} compares different regularization techniques with this network and reports test accuracy of $91.87\%$ with their probabilistic network and $90.88\%$ with dropout but omits BN. \cref{fig:BN-regularizes} shows how well BN generalizes for this problem when applied to exactly the {\em same} network. It easily achieves validation accuracy $93\%$, being significantly better than the dedicated regularization techniques proposed in~\cite{Gast18}. It appears that BN is a very powerful regularization method. The goal of this work is to try to understand and exploit the respective mechanism. Towards this end we identify two components: one is a non-linear reparametrization of the model that preconditions gradient descent and the other is stochasticity.

The reparametrization may be as well achieved by other normalization techniques such as {\em weight normalization}~\cite{Salimans2016WeightNA} and {\em analytic normalization}~\cite{shekhovtsov-18-norm} amongst others~\cite{Ba-2016-Layer-Norm,ArpitZKG16}. 
The advantage of these methods is that they are deterministic and thus do not rely on batch statistics, often require less computation overhead, are continuously differentiable~\cite{shekhovtsov-18-norm} and can be applied more flexibly, \eg to cases with a small batch size or recurrent neural networks.
Unfortunately, these methods, while improving on the training loss, do not generalize as good as BN, which was observed experimentally in~\cite{Gitman-17,shekhovtsov-18-norm}. We therefore look at further aspects of BN that could explain its regularization.

\begin{figure}[!t]
\centering
\setlength{\tabcolsep}{2pt}
\resizebox{\linewidth}{!}{
\begin{tabular}{cc}
\ \ \ \ \footnotesize{Training Loss} & \ \ \ \ \footnotesize Validation Loss \\
\includegraphics[width=0.47\linewidth]{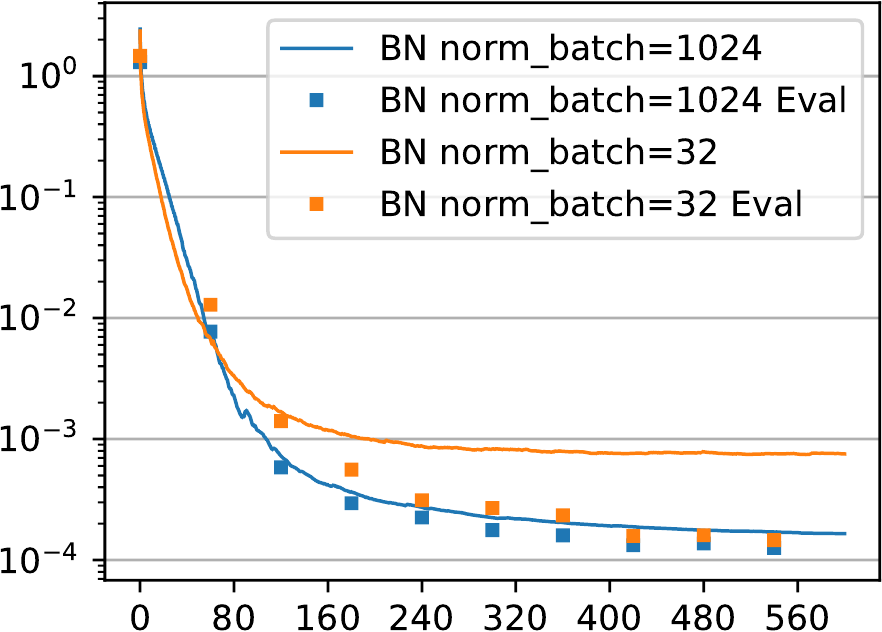}&%
\includegraphics[width=0.5\linewidth]{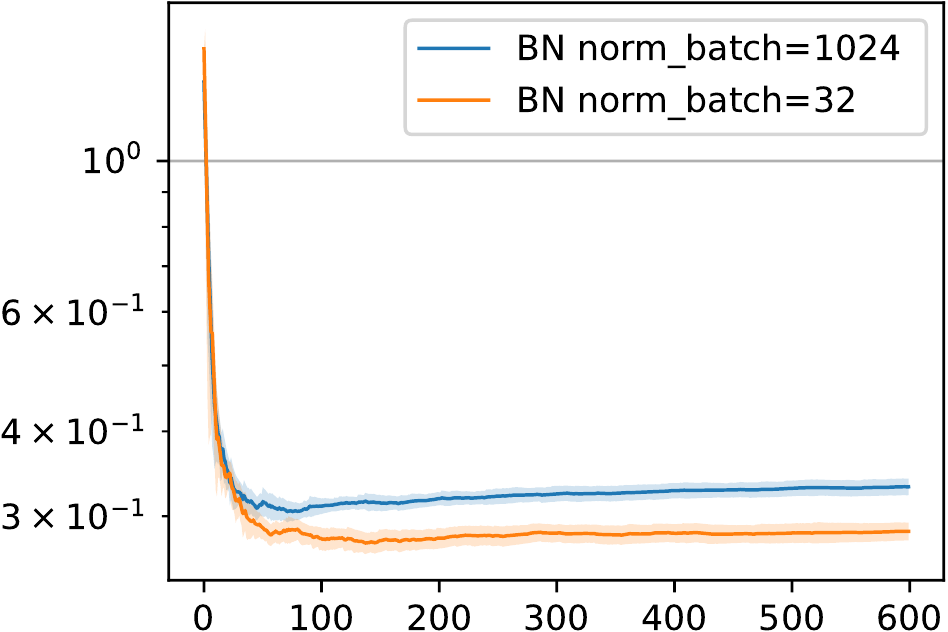}\\
\end{tabular}
}
\caption{The effect of randomness in BN. The training is performed with the same parameters. In the experiment with normalization batch 1024, BN layers use statistics of the data consisting of the training batch size 32 and 992 other samples chosen randomly at each step of SGD, all dependencies contribute to the total derivative. We observe that the gap between training and evaluation modes becomes smaller (as expected) but also that the validation loss increases. The validation accuracies are $93.0\pm 0.3$ and $92.4 \pm 0.3$, respectively.
\label{fig:BN-batch}}
\end{figure}

\citet{IoffeS15} suggest that the regularization effect of BN is related to the randomness of batch-normalization statistics, which is due to random forming of the batches. However, how and why this kind of randomness works remained unclear. Recent works demonstrated that this randomness can be reproduced~\cite{Teye-18} or simulated~\cite{Atanov-18} at test time to obtain useful uncertainty estimates. We investigate the effect of randomness of BN on training. For this purpose we first design an experiment in which the training procedure is kept exactly the same (including the learning rate) but the normalization statistics in BN layers are computed over a larger random subset of training data, the {\em normalization batch}. Note that the training batch size, for which the loss is computed is fixed to 32 throughout the paper. Changing this value significantly impacts the performance of SGD and would compromise the comparison of normalization techniques. The results shown in~\cref{fig:BN-batch} confirm that using larger normalization batches (1024) decreases the stochasticity of BN (the expected training loss gets closer to the evaluation model training loss) but, at the same time, its validation loss and accuracy get worse. The effect is not very strong, possibly due to the fact that BN in convolutional networks performs spatial averaging, significantly reducing the variance in most layers. Yet modeling and explaining it statistically allows to understand the connection to Bayesian learning and to apply such regularization with deterministic normalization techniques. Decoupling this regularization from the batch size and the spatial sizes of the layers and learning the appropriate amount of randomness allows to significantly reduce overfitting and to predict a better calibrated uncertainty at test time.

%
\subsection{Contribution}
We begin with the observation that BN and the deterministic normalization techniques rely on the same reparametrization of the weights and biases. In BN the node statistics taken over batches are used for normalization and followed by new affine parameters (scale and bias). We measure how random BN is, depending on the batch size and spatial dimensions of the network. We propose the view that BN can be represented as a noise-free normalization followed by a stochastic scale and bias. Next, we verify the hypothesis that such noises are useful for the two considered deterministic normalization techniques: weight normalization~\cite{Salimans2016WeightNA} and analytic normalization~\cite{shekhovtsov-18-norm}.
Furthermore, the view of stochastic scale and bias allows to connect BN to variational Bayesian inference~\cite{Graves-11} over these parameters and to variational dropout~\cite{Kingma-15-dropout}.
We test the complete Bayesian learning approach that learns noise variances of scale parameters and show that combining it with deterministic normalizations allows to achieve significant improvements in validation accuracy and loss. The results on the test set, on which we did not perform any parameter or model selection, confirm the findings.
\subsection{Related Work}
There are several closely related works concurrent with this submission~\cite{Santurkar-18,Teye-18,Atanov-18,XiangLi-18-BN}.
Work~\cite{Santurkar-18} argues that BN improves generalization because it leads to a smoother objective function,
the authors of \cite{XiangLi-18-BN} study the question why BN is often found incompatible with dropout,
and works~\cite{Teye-18, Atanov-18} observe that randomness in batch normalization can be linked to optimizing a  lower bound on the expected data likelihood~\cite{Atanov-18} and to variational Bayesian learning~\cite{Teye-18}. However, these works focus on estimating the uncertainty of outputs in models that have been already trained using BN. They do not make any proposals concerning the learning methods. The derivation and approximations made in~\cite{Teye-18} to establish a link to Bayesian learning is different from ours and, as we argue below, in fact gives poor recommendations regarding such learning. Overall, we remark that a better understanding of the success of BN is a topic of high interest.

The improved methods that we propose are also closely related to {\em variational drop\-out}~\cite{Kingma-15-dropout} as discussed below. We give a new interpretation to variational dropout and apply it in combination with normalization techniques.

\subsection{Background}
Let $X = w\T \hat X + a$ be an output of a single neuron in a linear layer. Batch normalization introduced by~\citet{IoffeS15} is applied after a linear layer before the non-linearity and has different training and test-time forms:
\begin{equation}\label{BN}
\begin{aligned}
&\mbox{training mode: \ \ } X' = \frac{X - M}{S}, \ \ \ \ \ \  && \mbox{evaluation mode:\ \ } X'' = \frac{X - \mu}{\sigma},
\end{aligned}
\end{equation}
where $(M, S^2)$ are the mean and variance statistics for a batch and $\mu$ and $\sigma^2$ are such statistics over the whole training distribution (in practice estimated with running averages during the training).
The normalized output $X'$ is invariant to the bias $a$ and to the scaling of the weight vector $w$, \ie, it projects out two degrees of freedom. They are then reintroduced after the normalization by an additional affine transformation $\tilde X = X' s + b$ with free parameters $s$ and $b$, so that the final class of modeled functions stays unchanged. BN has the following useful properties.
\begin{itemize}
\item Initialization. When a BN layer is introduced in a network, $s$ and $b$ are initialized as $s=1$, $b=0$. This resets the initial scale and bias degrees of freedom and provides a new initialization point such that the output of the BN layer will initially have zero mean and unit variance for the first training batch. The non-linearity that follows BN, will be not saturated for a significant portion of the batch data and the training may start efficiently.
\item Reparametrization. BN combined with a subsequent affine layer can be viewed as a non-linear reparametrization of the network. It was noted in~\cite{Salimans2016WeightNA} that such reparametrizations change the relative scales of the coordinates, which is equivalent to a preconditioning of the gradient (applying an adaptive linear transform to the gradient before each step)~\cite[\protect\parSym 8.7]{Luenberger:2015:LNP}.
\end{itemize}
Let us also note that a common explanation of BN as reducing the {\em internal covariate shift}~\cite{IoffeS15} was recently studied and found to be not supported by experiments~\cite{Santurkar-18}.

We will also consider {\em deterministic} normalization techniques, that do not depend on the selection of random batches: {\em  weight normalization} (WN)~\cite{Salimans2016WeightNA} and {\em analytic normalization}~\cite{shekhovtsov-18-norm}. We write all the discussed normalizations in the form
\begin{align}\label{normalization-general}
\tilde X = \frac{\w\T \hat X - \hat \mu(w)}{ \hat \sigma(w)} s + b,
\end{align}
where $\hat \mu(w)$ and $\hat \sigma(w)$ are different per method. For BN, they are the batch mean and standard deviation and depend on the batch as well as parameters of all preceding layers. For WN, $\hat \mu(w) = 0$ and $\hat \sigma(w) = \norma{w}$. It does the minimum to normalize the distribution, namely if $\hat X$ was a vector normalized to zero mean and unit variance then so is $w \hat X/\norma{w}$. However, if the assumption does not hold (due to the preceding non-linearities and scale-bias transforms), weight normalization cannot account for this.

For analytic normalization $\hat \mu(w)$ and $\hat \sigma(w)$ are the approximate statistics of $w \hat X$, computed by propagating means and variances of the training set through all the network layers. Thus they depend on the network parameters and the statistics of the whole dataset.
All three methods satisfy the following 1-homogeneity properties: $\hat \mu(\gamma w) = \gamma \hat \mu(w)$, $\hat \sigma(\gamma w) = |\gamma| \hat \sigma(w)$, which imply that~\eqref{normalization-general} is invariant to the scale of $w$ in all three methods.
\section{Importance of Reparametrization}\label{sec:reparam}
\begin{figure}[t]
\centering
\setlength{\tabcolsep}{2pt}
\resizebox{\linewidth}{!}{
\begin{tabular}{ccc}
\footnotesize \ \ \ \ BN & \footnotesize \ \ \ \ Weight Norm & \footnotesize \ \ \ \ Analytic Norm\\
\includegraphics[width=0.33\linewidth]{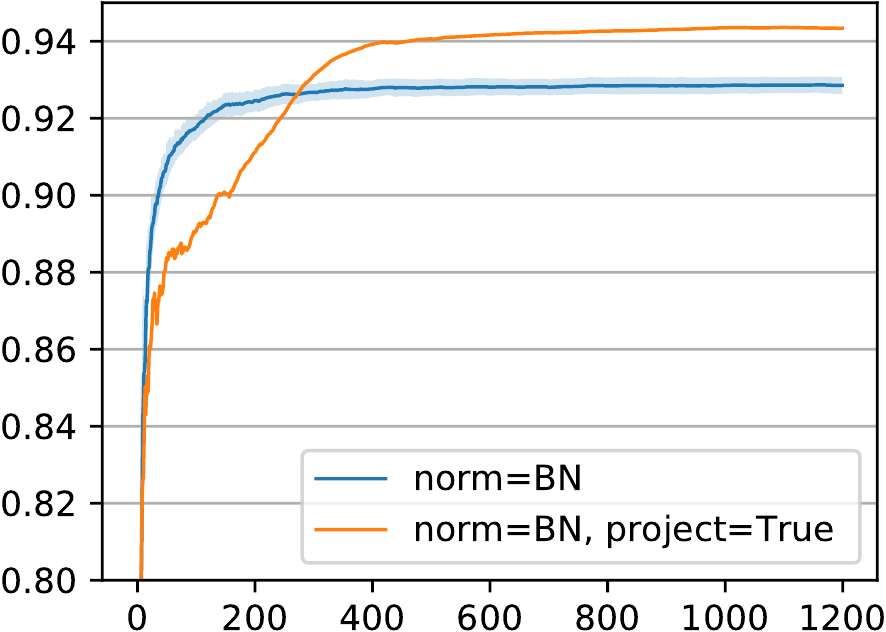}&%
\includegraphics[width=0.33\linewidth]{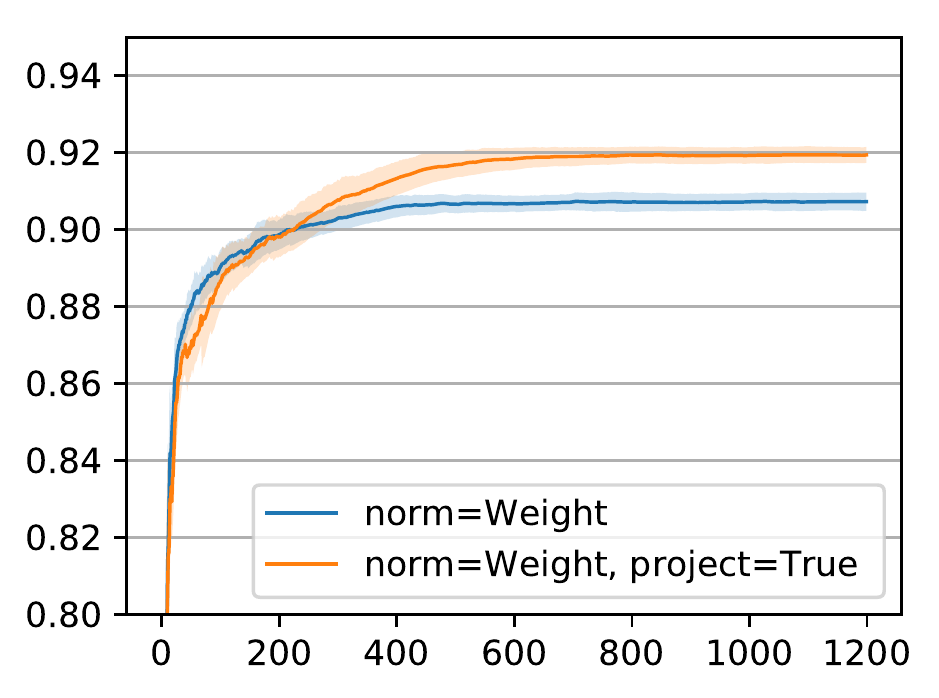}&%
\includegraphics[width=0.33\linewidth]{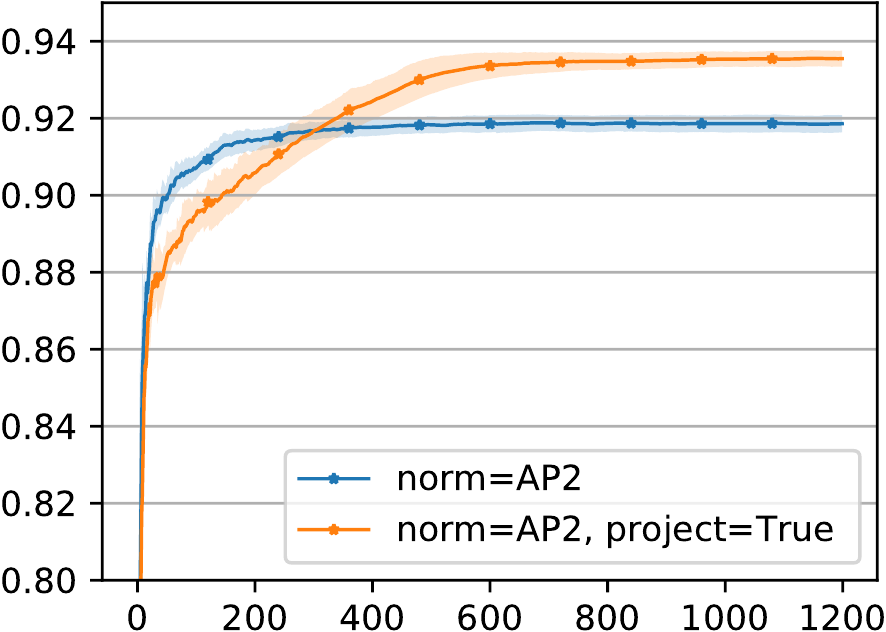}\\
\end{tabular}
}
\caption{Validation accuracies obtained by the normalization techniques with and without projection on constraints $\norma{w} = 1$ for all scaling invariant weights. Validation losses at the final point are improved as well as follows: Batch Norm: 0.31 $\rightarrow$ 0.22, Weigh Norm: 0.9 $\rightarrow$ 0.75, Analytic Norm:  0.7 $\rightarrow$  0.45.
\label{fig:project}}
\end{figure}
The original work~\cite{IoffeS15} recommended to use additional regularization $\lambda \norma{w}$ with {\em weight decay} $\lambda$. However, when used together with normalization, it leads to the learning problem of the form $\min_{w} f(w/\hat \sigma(w)) + \lambda \norma{w}$ (for clarity, we restrict the optimized parameters to the weight vector $w$ of a single neuron).  
This problem is ill-posed: it has no minimizer because decreasing the norm of $w$ is always a descent direction and at $\norma{w} = 0$ the function is undefined.

Many subsequent works nevertheless follow this recommendation, \eg~\cite{Salimans2016WeightNA,Gitman-17}. We instead propose to keep the constraint $\norma{w} = 1$ by projecting onto it after each gradient descent step. This avoids the possible instability of the optimization. Moreover, we found it to improve the learning results with all normalizations as shown by the following experiment. We compare learning with and without projection onto the constraint $\norma{w} = 1$ (no weight decay in both cases). 
The objective is invariant to $\norma{w}$, however, the optimization is not. 
~\cref{fig:project} shows experimental comparison for three normalization methods with or without projecting on the constrain $\norma{w} = 1$. It appears that projecting on the constraint has a significant impact on the validation performance. Notice that~\cite{Salimans2016WeightNA} propose quite the opposite with weight normalization: to allow $\norma{w}$ vary freely. Their explanation is as follows. The gradient of the reparametrized objective $f$ with respect to $w$ is given by
\begin{equation}
 \nabla_w f(w) = \frac{1}{\norma{w}} g_{\bot},
 \end{equation}
where $g$ is the gradient w.r.t. normalized weight $v = w/\norma{w}$ and $g_{\bot}$ denotes the components of $g$ orthogonal to $w$.
Thus, the gradient steps are always orthogonal to the weight vector $w$ and would progressively increase its norm. In its turn, the magnitude of the gradient in $w$ decreases with the increase of $\norma{w}$ and therefore smaller steps are made for larger $\norma{w}$. \cite{Salimans2016WeightNA} argues (theoretically and experimentally) that this is useful for optimization, automatically tuning the learning rate for $w$. We observed that in small problems, the norm $\norma{w}$ does not grow significantly, in which case there is no difference between projecting and not projecting. However, the experiments in~\cref{fig:project} show that in larger problems optimized with SGD there is a difference: allowing $\norma{w}$ to be free leads to smaller steps in $\norma{w}$ and to a worse accuracy in a longer run.

\section{Importance of Stochasticity}
It has been noted in~\cite{IoffeS15} that BN provides similar regularization benefits as dropout, since the activations observed for a particular training example are affected by the random selection of examples in the same mini-batch. In CNNs, statistics $M$ and $S^2$ in~\eqref{BN} are the sample mean and sample variance over the {\em  batch and spatial} dimensions:
\begin{align}
M = \frac{1}{n} \sum_{i=1}^{k}\sum_{j=1}^z X_{i,j}, \ \ \ \ \ \ S = \frac{1}{n} \sum_{i=1}^{k}\sum_{j=1}^z (X_{i,j} - M)^2,
\end{align}
where $k$ is the batch size, $z$ is the spatial size (we represent the spatial dimensions by a 1D index), $X_{i,j}$ is a response for sample $i$ at spatial location $j$ and $n = kz$. Because $M$ and $S$ depend on a random sample, for a given input $X=x$ the training-time BN output $X'$ can be considered as a random estimator of the test-time BN $x'' = (x-\mu)/\sigma$.
\subsection{Model of BN Stochasticity}
In this section we propose a simplified model of BN stochasticity replacing the randomness of batches by independent noises with known distributions. Despite the simplifying assumptions, this model allows to predict BN statistics and general dependencies such as the dependence on the batch size, which we then check experimentally.

For this theoretical derivation we will assume that the distribution of network activations $X$ over the full dataset is approximately normal with statistics $(\mu, \sigma^2)$. This assumption seems appropriate because the sample is taken over the whole dataset and also over multiple spatial locations in a CNN. We will also assume that the activations $X_{i,j}$ for different training inputs $i$ and different spatial coordinates $j$ are \iid. This assumption is a weaker one as we will see below. We can write the train-time BN as 
\begin{align}\label{BN-decomposed}
\frac{x - M}{S} = \Big(\frac{x - \mu}{\sigma} + \frac{\mu - M}{\sigma} \Big)\frac{\sigma}{S},
\end{align}
\ie expressing the output of the batch normalization through the exact normalization $(x - \mu)/\sigma$ (\cf test-time BN) and some corrections on top of it. Using the above independence assumptions, $M$ is a random variable distributed as $\N(\mu, {\textstyle \frac{1}{n}}\sigma^2)$. It follows that
\begin{align}
\frac{\mu - M}{\sigma} \sim \frac{1}{\sqrt{n}} \N(0,1) , \hspace{.5em} 
\frac{S^2}{\sigma^2} \sim \frac{1}{n}\chi^2_{n-1} \text{\hspace{.5em}and\hspace{.5em}} \frac{\sigma}{S} \sim \sqrt{n} \chi^{-1}_{n-1},
\end{align}
where $\chi^{2}$ is chi-squared distribution and $\chi^{-1}$ is the inverse chi distribution\footnote{
Using well known results for the distribution of the sample mean and variance of normally distributed variables. The inverse chi distribution is the distribution of $1/S$ when $S^2$ has a chi squared distribution~\cite{lee2012bayesian}.}.
The expression in~\eqref{BN-decomposed} has therefore the same distribution as
\begin{align}\label{BN model}
\Big(\frac{x - \mu}{\sigma} + V \Big) U,
\end{align}
if $V \sim \frac{1}{\sqrt{n}} \N(0,1)$ and $U\sim \sqrt{n} \chi^{-1}_{n-1}$ are independent \rv with known distributions (\ie, not depending on the network parameters).

\begin{figure}[t]
\centering
\includegraphics[height=0.21\linewidth]{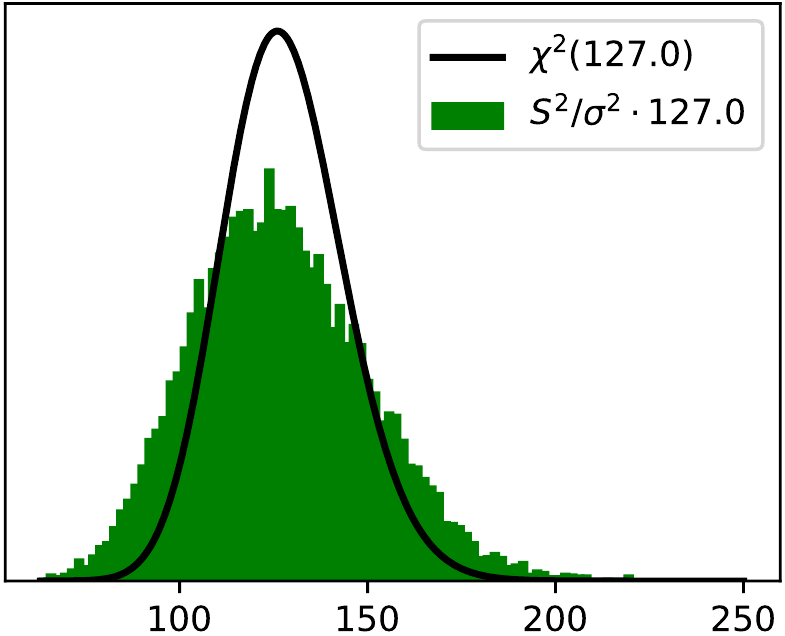}
\includegraphics[height=0.21\linewidth]{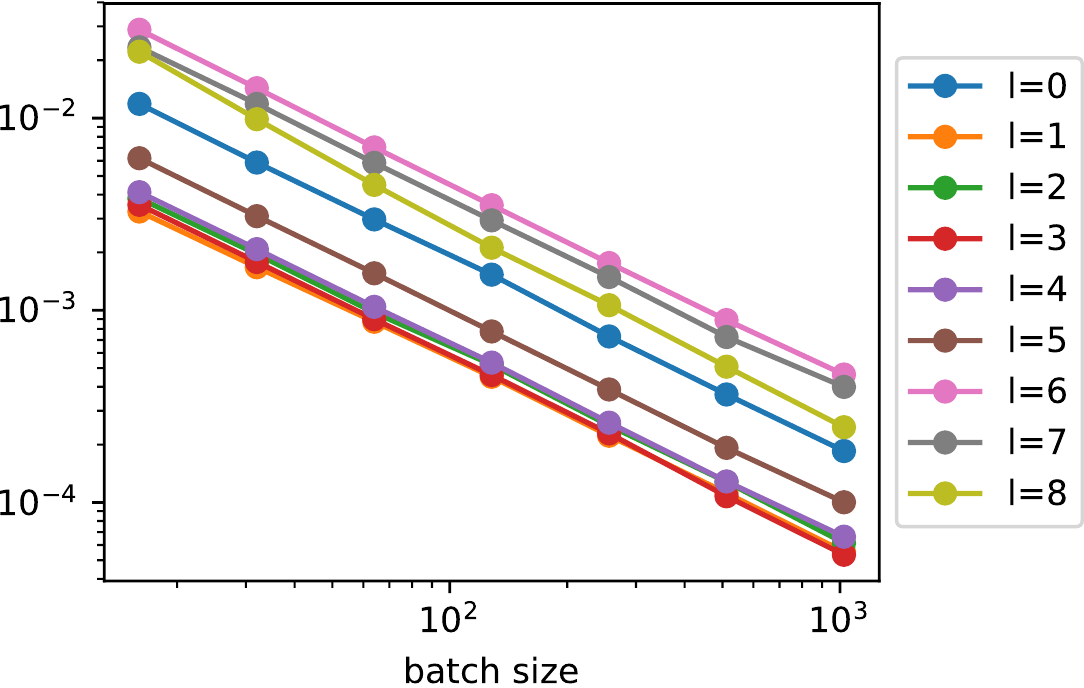}
\includegraphics[height=0.21\linewidth]{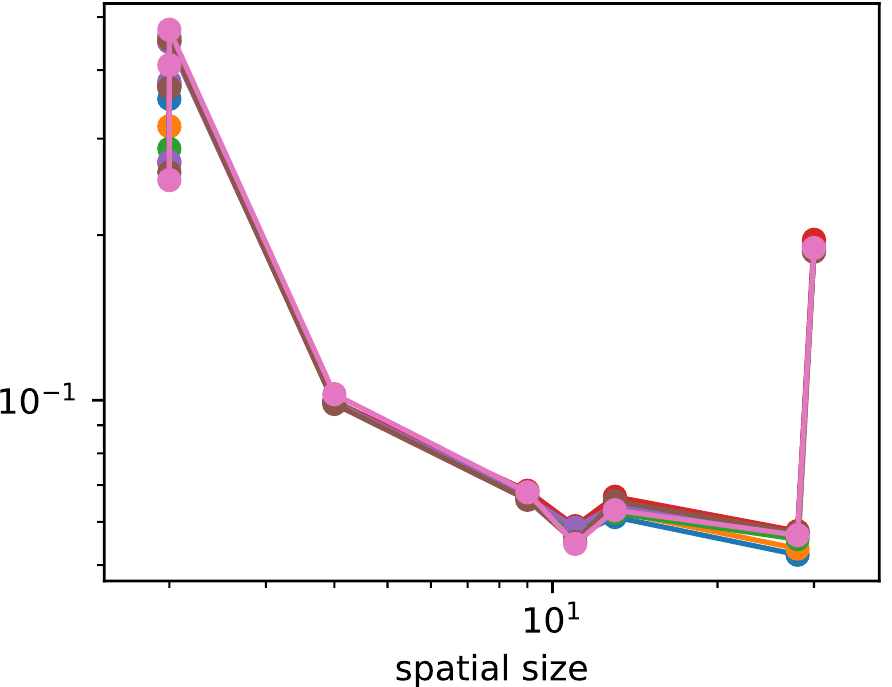}\\
\caption{Verification of BN statistics in a learned network.
{\em Left:} Observed distribution of $(n-1) S^2/\sigma^2$ in a unit in layer 6 of size $z = 2\times2$, batch size $k=32$, versus the model $\chi^2_{n-1}$.
{\em Middle:} measured variance of BN output versus batch size in different layers. The slope $-1$ in this log-log scale confirms $1/k$ dependence. {\em Right:} measured variance of BN output multiplied by $k$ versus the spatial size of a layer.
Here the slope is only approximately $-1$, due to spatial correlations, especially in the input layer (largest size).
\label{fig:BN-stochasticity}}
\end{figure}

We verify this model experimentally as follows. In a given (trained) network we draw random batches as during learning, propagate them through the network with BN layers and collect samples of $(\mu - M)/\sigma$ and $\sigma/S$ in each layer. We repeat this for several batch sizes. In \cref{fig:BN-stochasticity} left we see that the model of $\S^2/\sigma^2$ holds rather well: the real statistics is close to the theoretical prediction. Furthermore, the model predicts that the variance of BN (namely of the expression $(x-M/S)$) decreases as $1/(kz)$ with the batch size $k$ and spatial size $z$ of the current layer. \cref{fig:BN-stochasticity} middle clearly confirms the dependence on the batch size $k$  in all layers. The dependence on the spatial size $z$ (\cref{fig:BN-stochasticity} right) is not so precise, as the inputs are de facto spatially correlated, which we have ignored.

Concurrently to this work, the authors of~\cite{Atanov-18} have proposed a similar model for BN stochasticity and demonstrated that the distributions of $U$ and $V$ can be used at test time for improving the test data likelihoods and out-of-domain uncertainties. However, they did not explore using this model during the learning.
\subsection{Regularizing Like BN}\label{sec:reg-like}
\begin{figure}[t]
\centering
\setlength{\tabcolsep}{2pt}
\resizebox{\linewidth}{!}{
\begin{tabular}{rr}
\footnotesize \ \ \ \ Weight Normalization & \footnotesize \ \ \ \ Analytic Normalization \\
\includegraphics[width=0.47\linewidth]{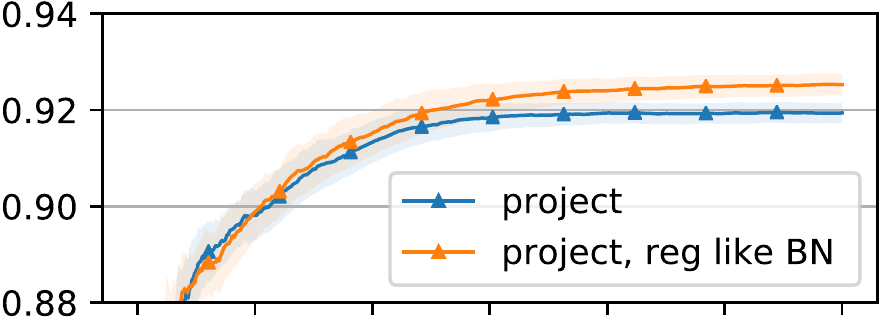}&%
\includegraphics[width=0.47\linewidth]{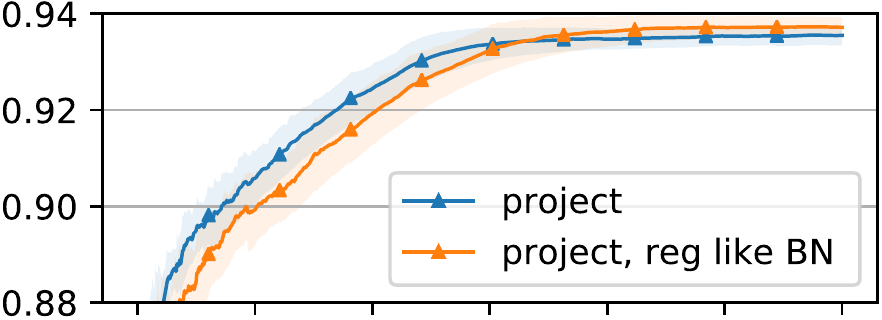}\\
\includegraphics[width=0.5\linewidth]{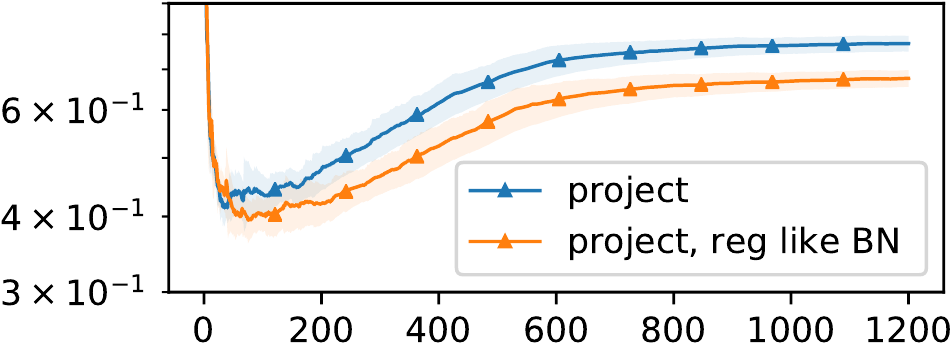}&%
\includegraphics[width=0.5\linewidth]{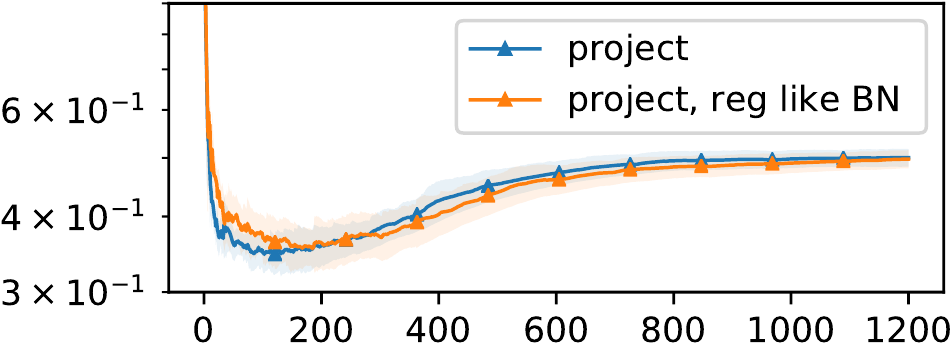}\\
\end{tabular}
}
\caption{Deterministic normalization techniques with noises like in BN. The noises appear to improve validation accuracy ({\em top}) and validation loss ({\em bottom}) noticeable for weight normalization and insignificantly for analytic normalization.
\label{fig:reg_like_BN}
}
\end{figure}

We now perform the following experiment. We measure the variances $\sigma^2_{V}$ and $\sigma^2_{U}$ of the random variables $V = (\mu - M)/\sigma$ and $U = \sigma/S$ in a network trained with BN. For example, the average standard deviation of the multiplicative noise, $\sigma_U$, in the consecutive layers was (0.05, 0.03, 0.026, 0.023, 0.02, 0.026, 0.041, 0.045, 0.071).
We then retrain the network with stochastic normalization using the expression~\eqref{BN model}, in which $\frac{x-\mu}\sigma$ is replaced with a deterministic method (either weight or the analytic normalization) and noises $V, U$ in~\eqref{BN model} are distributed as $\N(0, \sigma^2_{V})$ and $\N(1, \sigma^2_{U})$. It is important to note that the noises $V$, $U$ are {\em spatially correlated} as are the original quantities they approximate.
This may seem unnecessary, however we will see in the next section that these correlated activation noises can be reinterpreted as parameter noises and are closely related to Bayesian learning.
In~\cref{fig:reg_like_BN} we compare training using noise-free normalizations and noisy ones. The results indicate that injecting noises to deterministic normalization techniques does regularize the training but the amount of noise can be still increased to make it more efficient. In the next section we consider learning the noise values instead of picking them by hand.

We argue that the combination of noise following the normalization is particularly meaningful. The base network, without normalizations is equivariant to the global scale of weights. To see this, note that linear layers and ReLU functions are 1-homogeneous: scaling the input by $\gamma$ will scale the output by $\gamma$. Consider an additive noise $\xi$ injected in front of non-linearities as was proposed \eg in~\cite{Gast18}:
\begin{align}
\dots W^{k+1} \mbox{ReLU} (W^{k} x + b^k + \xi ) + b^{k+1} \dots,
\end{align}
where $\xi$ has a fixed distribution such as $\N(0, 0.01)$~\cite{Gast18}. Then, scaling $W^{k}$ and $b^k$ by $\gamma > 1$ and scaling $W^{k+1}$ by $1/\gamma$ allows the model to increase the signal-to-noise ratio and to suppress the noise completely. 
In contrast, when noises are injected after a normalization layer, as in~\eqref{BN-decomposed}, the average signal to noise ratio is kept fixed. 

\subsection{BN as Bayesian Learning}
Let $D = ((x^t, y^t) \mid t = 1,\dots |D|)$ be our training data. In Bayesian framework, given a prior distribution of parameters $p(\theta)$, first the posterior parameter distribution $p(\theta | D)$ given the data is found, then for prediction we marginalize over parameters:
\begin{align}\label{Bayes-posterior}
p(y \mid x, D) = \int p(y \mid x, \theta) p(\theta \mid D) d \theta ,
\end{align}
A practical Bayesian approach is possible by using a variational approximation to the parameter posterior, as was proposed for neural networks by~\cite{Graves-11}.
It consists in approximating the posterior $p(\theta \mid D)$ by a simple distribution $q(\theta)$ from a parametric family with parameters $\phi$, for example, a diagonal multivariate normal distribution $\N(\hat \theta, \sigma^2)$ with $\phi = (\hat \theta, \sigma)$. 
The approximate Bayesian posterior~\eqref{Bayes-posterior} becomes
\begin{align}\label{Bayes-posterior-approx}
p(y \mid x, D) \approx \int p(y \mid x, \theta) q(\theta) d \theta,
\end{align}
which allows at least for the Monte Carlo (MC) approximation. 
The distribution $q$ is found by minimizing the KL-divergence between $q(\theta)$ and $p(\theta \mid D)$
\begin{align}\label{variational Bayes}
{\rm KL}(q(\theta) \| p(\theta \mid D)) = \E\limits_{\theta \sim q}\Big[{-}\hskip-3pt\sum_{t}\log p(y^t | x^t, \theta)\Big]  + {\rm KL}(q(\theta) \| p(\theta)) + const
\end{align}
in parameters $\phi$. When $q$ is chosen to be the delta-function at $\hat \theta$ (\ie a point estimate of $\theta$) and the prior $p(\theta)$ as $\N(0,1/\lambda I)$,
formulation~\eqref{variational Bayes} recovers the conventional maximum likelihood~\cite{Graves-11} regularized with $\lambda \|\hat\theta\|^2 / 2$. The first term in~\eqref{variational Bayes}, called the {\em data evidence} can be written as the joint expectation over parameters and data:
\begin{align}\label{data evidence}
|D| \E_{\larr{\theta \sim q\\ (x,y) \sim D}} \Big[ -\log p(y \mid x, \theta)\Big],
\end{align}
where $(x,y) \sim D$ denotes drawing $(x, y)$ from the training dataset uniformly. 
The gradient \wrt $\phi$ of the expectation~\eqref{data evidence} for normal distribution $q(\theta)$ (assuming $p(y \mid x, \theta)$ is differentiable a.e.) expresses as
\begin{align}\label{grad data evidence}
|D| \E_{\larr{\theta \sim q\\ (x,y) \sim D}}\Big[ -\nabla_\phi \log p(y \mid x, \theta)\Big].
\end{align}
Using the parametrization $\theta = \hat \theta + \sigma \xi$, $\xi \sim \N(0,1)$, gradient~\eqref{grad data evidence} simplifies to
\begin{align}
|D| \E_{\larr{\xi\sim \N(0,1) \\ (x,y) \sim D}}\Big[ -\nabla_\phi \log p(y \mid x, \hat \theta + \sigma \xi )\Big].
\end{align}
For a more general treatment of differentiating expectations see~\cite{Schulman-15}.
A stochastic gradient optimization method may use an unbiased estimate with mini-batches of size $M$:
\begin{align}\label{data-evidence-rep}
\frac{|D|}{M} \sum_{m=1}^{M} -\nabla_\phi \log p(y^m \mid x^m, \hat \theta + \sigma \xi).
\end{align}
This means that during learning we randomly perturb parameters for every input sample.
It becomes apparent that {\em any noises in the parameters during the standard maximum likelihood learning are closely related to the variational Bayesian learning}.

In order to connect BN to Bayesian learning, it remains to define the form of the approximating distribution $q$ that would correspond to the noisy model of BN with an affine transform as given by
\begin{align}\label{BN-affine}
\Big(\frac{w\T x-\mu}{\sigma} + V \Big) U s + b.
\end{align}
We reinterpret~\eqref{BN-affine} as a model with a stochastic affine transform defined by a stochastic scale $S$ and a stochastic bias $B$:
\begin{align}\label{BN-Affine}
\Big(\frac{w\T x-\mu}{\sigma} + B \Big) S.
\end{align}
We define the approximate posterior $q$ over parameters $w, S, B$ to be factorized as $q(w)q(S,B)$, where $q(w)$ is a delta distribution, \ie~a point estimate, and $q(S,B)$ is a coupled distribution over scale and bias defined as a distribution of a parametric mapping of independent random variables $U$ and $V$:
\begin{align}\label{BN-noise params}
S = U s,\ \ \ \ B = V + b/(Us).
\end{align}
We let the prior $p(S, V)$ to be defined through the prior on $U,V$ and the same mapping~\eqref{BN-noise params}. The invariance of KL divergence under parameter transformations allows us to write 
\begin{align}\label{BN-prior}
{\rm KL}(q(S,B) \| p(S,B)) = {\rm KL}(q(U) \| p(U)) + {\rm KL}(q(V) \| p(V)).
\end{align}
This completes the construction, which can now be summarized as the following proposition.
\begin{proposition}
Assuming that BN can be well approximated with a noisy normalization modeled by~\eqref{BN model}, the BN learning is equivalent to variational Bayesian learning~\cite{Graves-11} with stochastic scale-bias parameter distribution with fixed $q(U)$, $q(V)$ and prior on $w$ which is uniform on the sphere $\norma{w} = 1$ (for all normalized units).
\end{proposition}
Note, that the distribution of $q(S,B)$ still depends on two parameters and is optimized, but the KL divergence term vanishes.
In other words, BN, optimizes only the data evidence term, but does not have the KL prior (it is considered constant). Note also that by choosing a suitable prior, a regularization such as $\norma{b}^2$ can be derived as well.

Concurrently to this work,~\cite{Teye-18} proposed an explanation of BN as Bayesian learning with a different interpretation. They associate the stochasticity to noises in the linear transform parameters $w$ and build a sequence of approximations to justify the weight norm regularization $\norma{w}$ as the KL prior. While this matches the current practices of applying weight decay, it leads to the problem of regularizing the degrees of freedom to which the network is invariant, as we discussed in~\cref{sec:reparam}. It is therefore likely that some of the approximations made in~\cite{Teye-18} are too weak. Furthermore, the authors do not draw any applications of their model to experiments or learning methods.

Our interpretation is also only an approximation based on the simplified stochastic model of BN~\eqref{BN model}. However, one more argument in favor of the Bayesian learning view of BN is the following. It appears that from the initialization to convergence of BN learning, the standard deviations of $U$, $V$ are in fact changing, especially in the final layer, growing by a factor of up to 5. So BN appears to increase the regularization towards convergence, which is also the case when we optimize these distributions with Bayesian learning.

\subsection{Connection to Variational Dropout}
\citet{Kingma-15-dropout} proposed a related regularization method called {\em variational dropout}. More specifically, in the case of ReLU non-linearities and fully connected linear layers, the non-negative stochastic scaling $S$ applied to the output of a linear layer~\eqref{BN-Affine}, can be equivalently applied to the input of the subsequent linear layer, which then expresses as
\begin{align}\label{V-dropout}
\sum_j W_{ij} (x_j S_j) + b_i = \sum_j (W_{ij} S_j) x_j + b_i,
\end{align}
matching the {\em variational dropout with correlated white noise} \citep[sec. 3.2]{Kingma-15-dropout}. They consider approximate Gaussian posterior $q(S) \sim \N(1, \sigma^2)$, a log-uniform prior on $S$, given by $p(\log(|S|)) = const$ and no priors on $w$ and $b$ and apply the variational Bayesian learning to this model. 
This model is rather economical, in that an extra variance variable $\sigma^2$ is introduced per input channel and not per weight coefficient and performed better than the other studied variants of variational dropout~\cite{Kingma-15-dropout}.
Extending this model to the convolutional networks retains little similarity with the original dropout~\cite{srivastava14a}. The main difference being that the noise is applied to parameters rather than activations. See also~\cite{Gal-16-RNN} discussing variational dropout and such correlations in the context of RNNs.

\subsection{Normalization with Bayesian Learning}
We propose now how the model~\eqref{BN-Affine} can be applied with other normalization techniques. 
For simplicity, we report results with bias $b$ being deterministic, \ie consider the model
\begin{align}\label{Norm-Bayes}
\Big(\frac{w\T x-\mu}{\sigma} + b \Big) S.
\end{align}
We expect that the normalized output $(w\T x-\mu)/\sigma$ approximately has zero mean and unit variance over the dataset.
This allows to set reasonable prior for $S$. In our experiments we used $p(S) \sim \N(1, 10^2)$ (we do not expect scaling by a factor more than 10 in a layer, a rather permissive assumption) and no prior on $b$.
We then seek point estimates for $w$ and $b$ and a normal estimate of $S$ parametrized as $q(S) \sim \N(s, \sigma^2_S)$. A separate value of variance $\sigma^2_S$ may be learned per-channel or just one value may be learned per layer. In the former case, the learning has a freedom to chose high variances for some channels and in this way to make an efficient selection of model complexity.

The KL divergence on the scale parameter ${\rm KL}(q(S) \| p(S))$ with these choices is, up to constants, $-\log \sigma^2 + \frac{\sigma^2}{\sigma^2_0} + \frac{(s-1)^2}{\sigma^2_0}$.
We observe, that the most important term, the only one that pushes the variance $\sigma$ up and prevents overfitting is $-\log \sigma^2$. The very same term occurs in the KL divergence to the scale-uniform prior~\cite{Kingma-15-dropout}. The remaining terms balance how much of variance is large enough, \ie may only decrease the regularization strength and penalize large value of $s$.

We identified one technical problem with such KL divergences when used in stochastic gradient optimization: when $\sigma$ approaches zero, the derivative $-1/\sigma$ is unbounded. This may cause instability of SGD optimization with momentum. To address this issue we reparametrize $\sigma$ as a piece-wise function $\sigma = e^u$ if $u < 0$ and $\sigma = u + 1$ if $u\geq 0$. This makes sure that derivatives of both $\log \sigma$ and $\sigma$ are bounded. Note that a simpler parametrization $\sigma = e^u$ has quickly growing derivatives of the linear terms in $\sigma$ and that the data evidence as composition of log softmax and piecewise-linear layers is approximately linear in each variance $\sigma$ as seen from the parametrization~\eqref{data-evidence-rep}.
Note that using a sampling-based estimate of the KL divergence as in~\cite{Blundell-15} does not circumvent the problem because it contains exactly the same problematic term $-\log \sigma$ in every sample.

\begin{figure}[t]
\centering
\setlength{\tabcolsep}{2pt}
\resizebox{\linewidth}{!}{
\begin{tabular}{ccc}
\includegraphics[width=0.33\linewidth]{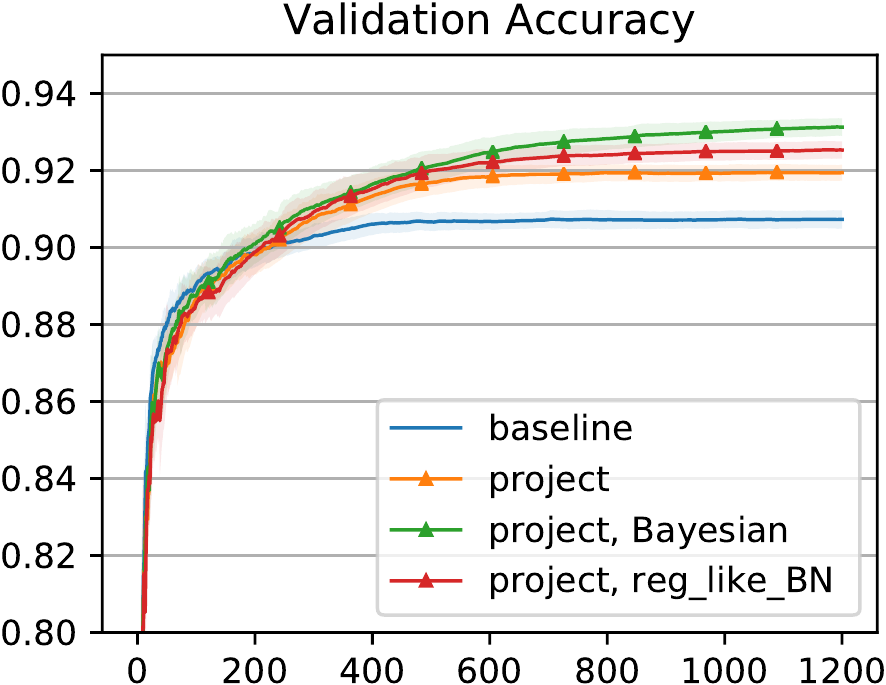}&%
\includegraphics[width=0.35\linewidth]{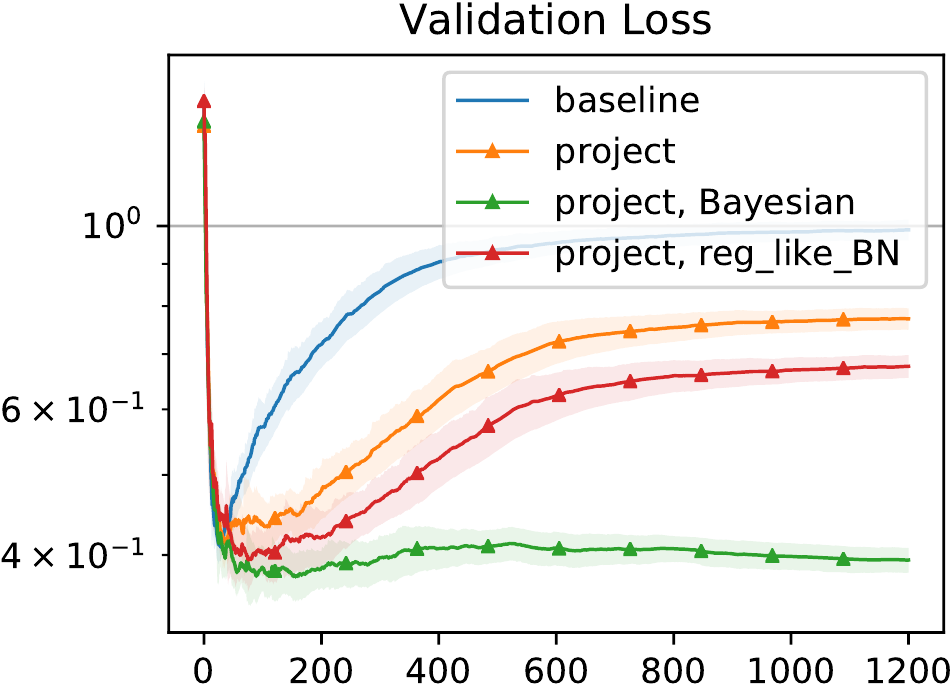}&%
\includegraphics[width=0.33\linewidth]{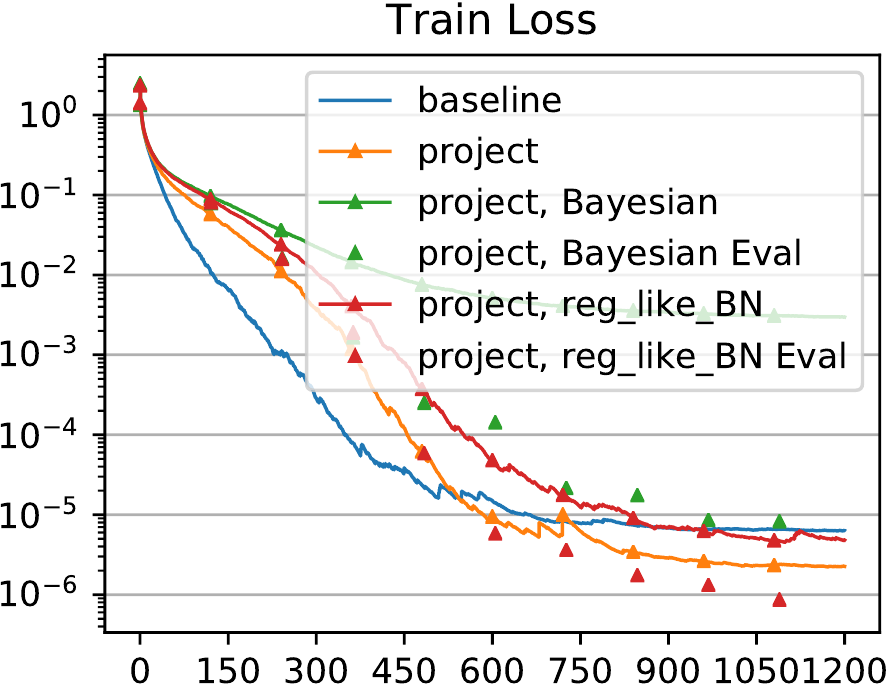}\\
\end{tabular}
}
\caption{All improvements to weight normalization. Bayesian Learning leads to significantly reduced validation loss and improved accuracy.
For Bayesian learning, the {\em training loss} shows as solid line the expected data evidence estimated as running mean during training and  triangles show the training loss with mean parameters. The {\em validation loss} plot uses mean parameters.
The MC estimate of the validation loss of the final model with 10 samples is $0.25$, significantly lower than the value $0.4$ in the plot when using mean parameters.
\label{fig:WN-Bayes}}
\end{figure}
\begin{figure}[t]
\centering
\setlength{\tabcolsep}{2pt}
\resizebox{\linewidth}{!}{
\begin{tabular}{ccc}
\includegraphics[width=0.33\linewidth]{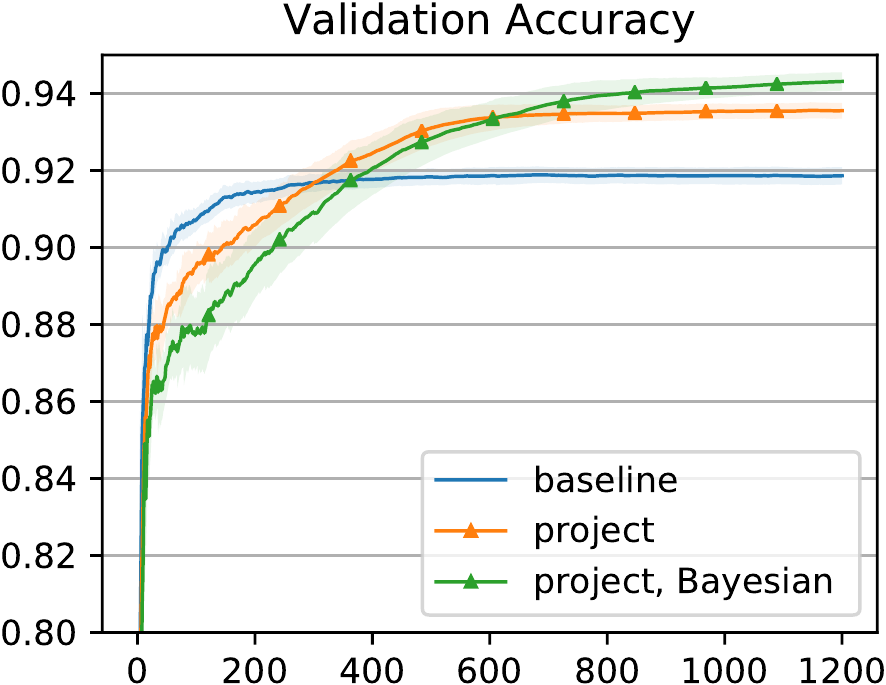}&%
\includegraphics[width=0.35\linewidth]{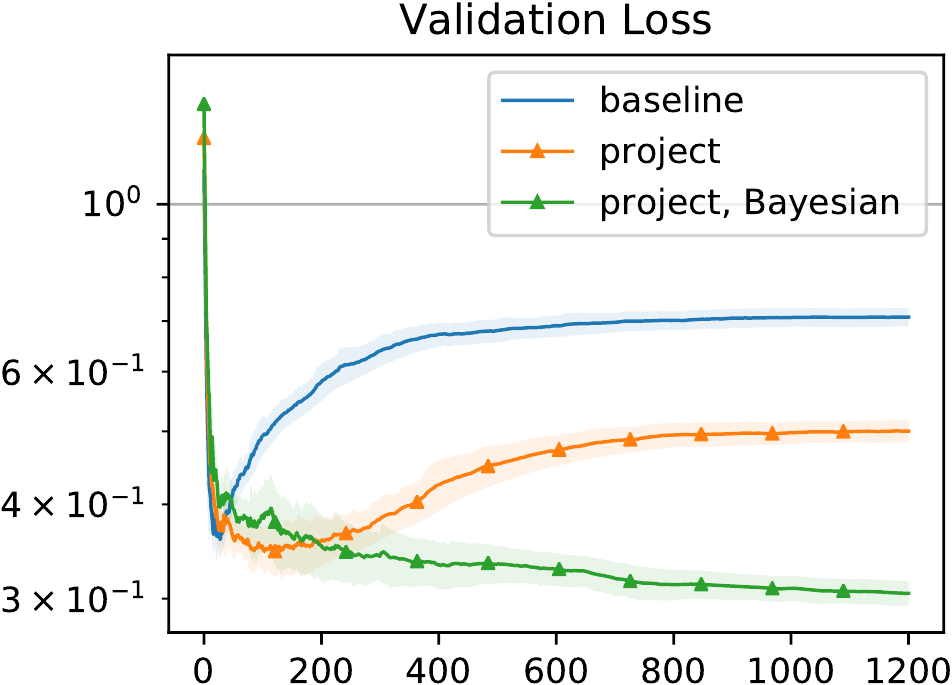}&%
\includegraphics[width=0.33\linewidth]{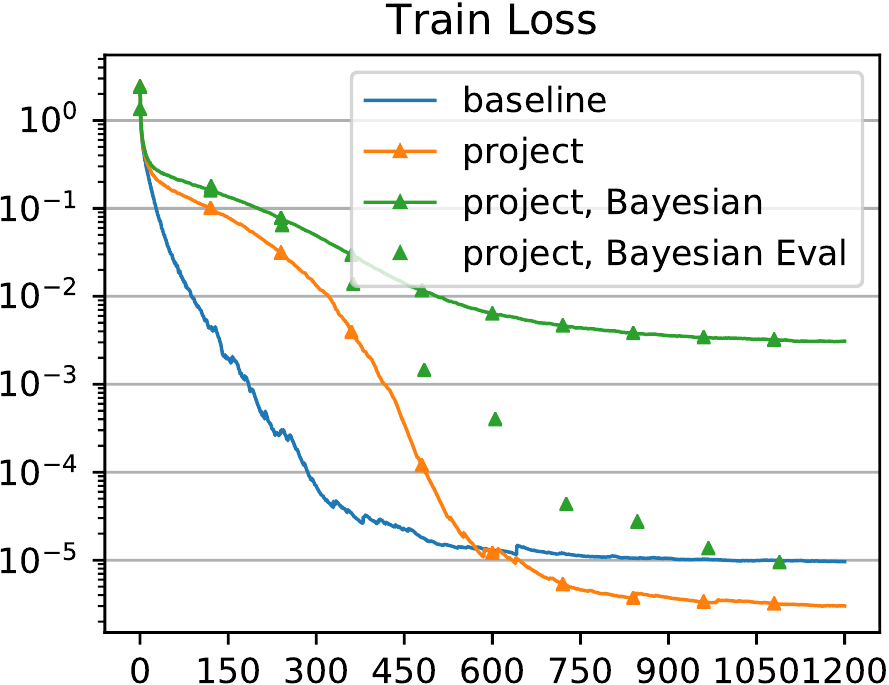}\\
\end{tabular}
}
\caption{Analytic Normalization with Bayesian learning in comparison with projection only and with the baseline version~\cite{shekhovtsov-18-norm}.
\label{fig:VN-Bayes}}
\end{figure}

\cref{fig:WN-Bayes,fig:VN-Bayes} show the results for this Bayesian learning model with weight and analytic normalization. In the evaluation mode we substitute mean values of the scales, $s$. In~\cref{tab:results-test} we also show results obtained by MC estimation of the posterior~\eqref{Bayes-posterior}.
It is seen that the validation accuracy of both normalizations is improved by the Bayesian learning. Its most significant impact is on the validation loss, which is particularly important for an accurate estimation of uncertainties for predictions made with the model. MC estimates of the validation losses, better approximating the Bayesian posterior~\eqref{Bayes-posterior-approx}, are yet significantly lower as seen in~\cref{tab:results-test}. It is interesting to inspect the learned noise values. For analytic normalization we obtained in the consecutive layers the average $\sigma_S/s$ = (0.39, 0.34, 0.18, 0.29, 0.27, 0.29, 0.31, 0.61, 0.024), corresponding to standard deviation of $U = S/s$. Compared the respective noises in BN~\cref{sec:reg-like}, these values are by up to an order of magnitude larger except in the last layer. The learned noises put more randomness after the input and in the penultimate layer that has 192 channels. The final linear transform to 10 channels followed by spatial pooling can indeed be expected to tolerate more noise.
\begin{table}[t]
\resizebox{0.8\linewidth}{!}{
\setlength{\tabcolsep}{3pt}
\begin{tabular}{|c|c|c|c|c|}
\hline
\multirow{2}{*}{Method} & \multirow{2}{*}{Test accuracy, \%} & \multicolumn{3}{c|}{Test negative log likelihood}\\
\cline{3-5}
& & Single-pass & MC-10 & MC-30 \\
\hline
No normalization & 90.7 & 1.45 & - &  - \\
\hline
Baseline BN & 92.7 & 0.34 & - &  - \\
\hline
BN with Projection & 94.1 & {\bf 0.29} & - &  - \\
\hline
Weight Normalization & 93.5 & 0.48 & 0.27 & 0.24 \\
\hline
Analytic Normalization & {\bf 94.4} & 0.38 & {\bf 0.22} & {\bf 0.20} \\ 
\hline
\multicolumn{5}{l}{Best previously published results with the same network}\\
\hline
Dropout~\cite{Gal-16} as reported in~\cite{Gast18} & 90.88 & - & - & 0.327\\
\hline
ProbOut~\cite{Gast18} & 91.9 & 0.37 & - & - \\
\hline
\multicolumn{5}{l}{Published results with other networks}\\
\cline{1-2}
ELU~\cite{ClevertUH15} & 93.5 & \multicolumn{3}{c}{}\\
\cline{1-2}
ResNet-110~\cite{HeZRS16} & 93.6 & \multicolumn{3}{c}{}\\
\cline{1-2}
Wide ResNet~\cite{Zagoruyko-16-WRN} (includes BN) & {\bf 96.0}  & \multicolumn{3}{c}{}\\
\cline{1-2}
\end{tabular}
}
\caption{Summary of results for the {\em test set} in CIFAR-10.
The test set does not contain augmentations and was not used in any way during training and parameter selection. Weight normalization and analytic normalization use projection and Bayesian learning. For comparison we also quote recently published results in~\cite{Gast18} for the very same network and state-of-the art results with more advanced networks. We did not run our method with these larger networks.
\label{tab:results-test}
}
\end{table}
\begin{figure}[!t]
\centering
\setlength{\tabcolsep}{2pt}
\resizebox{\linewidth}{!}{
\begin{tabular}{ccc}
\includegraphics[width=0.33\linewidth]{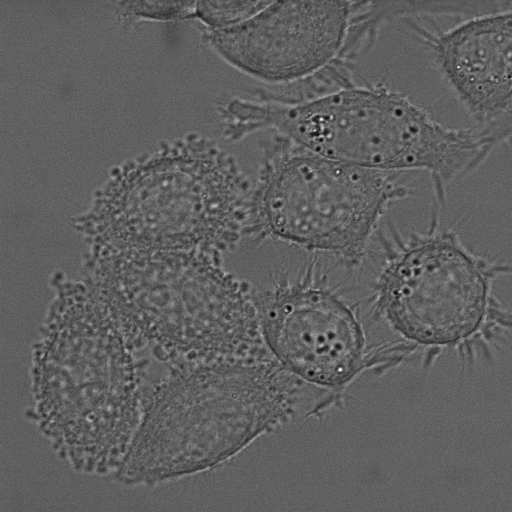}&%
\includegraphics[width=0.33\linewidth]{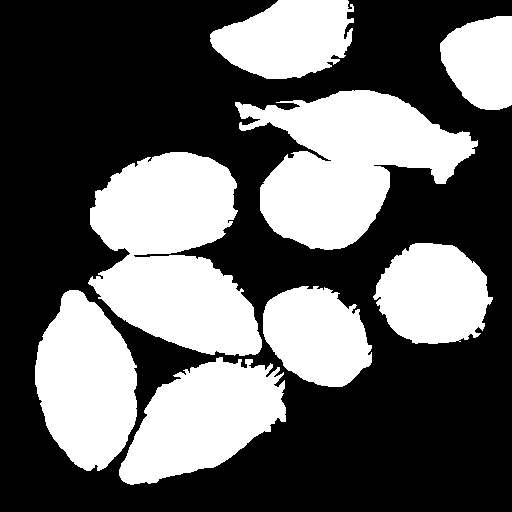}&%
\includegraphics[width=0.33\linewidth]{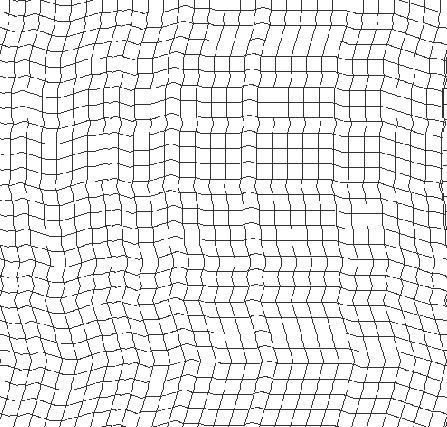}\\
\end{tabular}
}
\caption{{\em Left:} exemplar image from the dataset (512x512). {\em Middle:} exemplar training segmentation. {\em Right:} exemplar deformation applied to a grid image.
\label{fig:hela-data}}
\end{figure}
\begin{figure}[!t]
\centering
\setlength{\tabcolsep}{2pt}
\begin{tabular}{cc}
\ \ \ \ \footnotesize{Validation Loss} & \ \ \ \ \footnotesize Validation Accuracy \\
\includegraphics[width=0.45\linewidth]{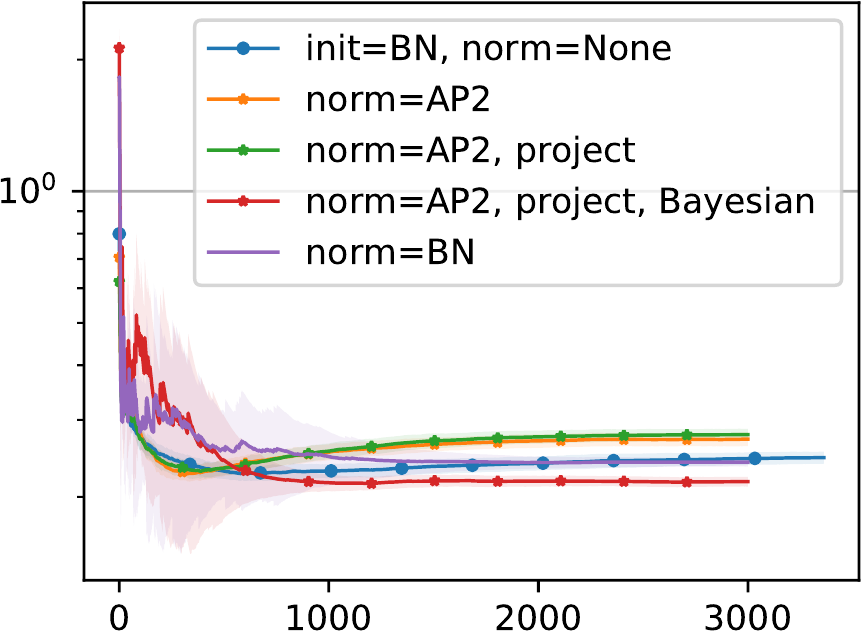}&%
\includegraphics[width=0.45\linewidth]{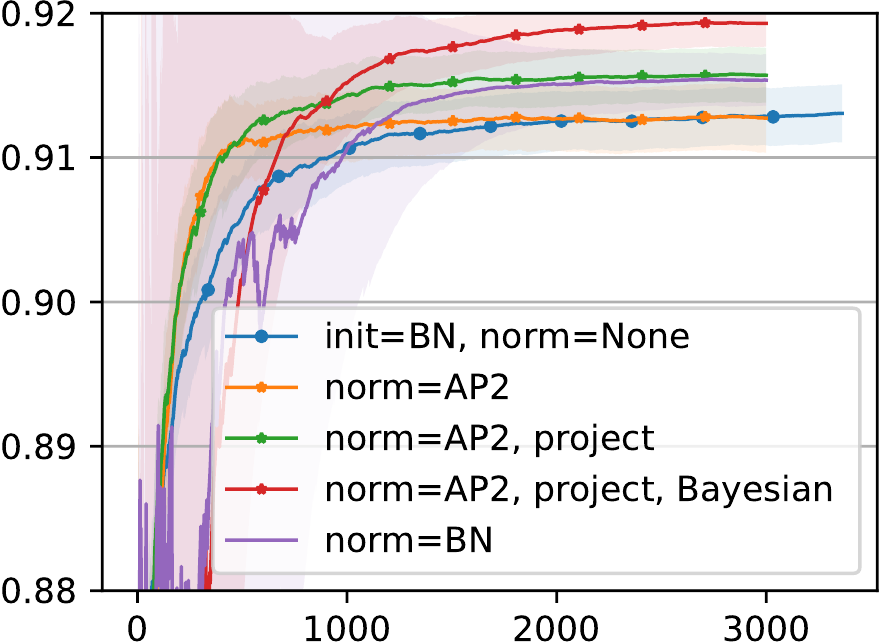}\\
\end{tabular}
\caption{Comparison of methods on the cell segmentation dataset DIC-HeLa~\cite{cell-tracking-algs} with a deep fully convolutional network (11 conv layers).
\label{fig:hela}}
\end{figure}
\section{Other Datasets}
To further verify applicability of our proposed improvements, we made an experiment with a quite different problem: segmentation of the dataset "DIC-HeLa" from the ISBI cell tracking challenge~\cite{cell-tracking-algs} illustrated in~\cref{fig:hela-data}. The difficulty of this dataset is that it contains only 20 fully annotated training images, and the cells are hard to segment. Even when using significant data augmentation by non-rigid transforms as illustrated in~\cref{fig:hela-data} right, there is still a gap between training and validation accuracy. The Bayesian learning approach in combination with analytic normalization gives a noticeable improvement as shown in~\cref{fig:hela}. Please refer to~\cref{sec:hela-supp} for details of this experiment. One of the open problems, is how to balance the prior KL divergence term, as in the case of a fully convolutional networks with data augmentation, we do not have a clear notion of a number of training examples. Results in~\cref{fig:hela} are obtained with KL factor $0.1$ per classified pixel (meaning the augmentation is worth 10 examples).
%
%
%
%
\section{Conclusion}
We have studied two possible causes for a good regularization of BN. The effect achieved due to the interplay of the introduced reparametrization and SGD appears to play the major role. We have improved this effect empirically by showing that performing the optimization in the normalized space improves generalization for all three investigated normalization methods. We have analyzed the question of how the randomness of batches helps BN. The effect was quantified, the randomness measured and modeled as injected noises. The interpretation as Bayesian learning is a plausible explanation for why it helps to improve generalization. We further showed that such regularization helps other normalization techniques to achieve similar performance. This allows to improve performance in scenarios when BN is not suitable and to learn the model randomness instead of fixing it by batch size and the architecture of the network. We found that variational Bayesian learning may occasionally diverge due to a delicate balance of the KL divergence prior. This question and the utility of the learned uncertainty are left for future work.
%
\section*{Acknowledgments}
A.S. has been supported by Czech Science Foundation grant 18-25383S and Toyota Motor Europe. B.F.~gratefully acknowledges support by the Czech OP
VVV project "Research Center for
Informatics" (CZ.02.1.01/0.0/0.0/16\_019/0000765). 


\renewcommand{\bibname}{\protect\leftline{References\vspace{-2ex}}}
\bibliographystyle{splncsnat}
\bibliography{../bib/strings,../bib/neuro-generative,../bib/our}

\begin{thebibliography}{26}
\providecommand{\natexlab}[1]{#1}
\providecommand{\url}[1]{\texttt{#1}}
\providecommand{\urlprefix}{}

\bibitem[{Arpit et~al.(2016)Arpit, Zhou, Kota, and Govindaraju}]{ArpitZKG16}
Arpit, D., Zhou, Y., Kota, B.U., Govindaraju, V.: Normalization propagation:
  {A} parametric technique for removing internal covariate shift in deep
  networks.
\newblock In: ICML. pp. 1168--1176 (2016)

\bibitem[{Atanov et~al.(2018)Atanov, Ashukha, Molchanov, Neklyudov, and
  Vetrov}]{Atanov-18}
Atanov, A., Ashukha, A., Molchanov, D., Neklyudov, K., Vetrov, D.: Uncertainty
  estimation via stochastic batch normalization.
\newblock In: ICLR Workshop track (2018)

\bibitem[{Blundell et~al.(2015)Blundell, Cornebise, Kavukcuoglu, and
  Wierstra}]{Blundell-15}
Blundell, C., Cornebise, J., Kavukcuoglu, K., Wierstra, D.: Weight uncertainty
  in neural networks.
\newblock In: ICML. pp. 1613--1622 (2015)

\bibitem[{Clevert et~al.(2016)Clevert, Unterthiner, and
  Hochreiter}]{ClevertUH15}
Clevert, D.A., Unterthiner, T., Hochreiter, S.: Fast and accurate deep network
  learning by exponential linear units ({ELUs}).
\newblock In: ICLR (2016)

\bibitem[{Gal and Ghahramani(2016{\natexlab{a}})}]{Gal-16}
Gal, Y., Ghahramani, Z.: Dropout as a {B}ayesian approximation: Representing
  model uncertainty in deep learning.
\newblock In: ICML. pp. 1050--1059 (2016{\natexlab{a}})

\bibitem[{Gal and Ghahramani(2016{\natexlab{b}})}]{Gal-16-RNN}
Gal, Y., Ghahramani, Z.: A theoretically grounded application of dropout in
  recurrent neural networks.
\newblock In: NIPS. pp. 1027--1035 (2016{\natexlab{b}})

\bibitem[{Gast and Roth(2018)}]{Gast18}
Gast, J., Roth, S.: Lightweight probabilistic deep networks.
\newblock In: CVPR (June 2018)

\bibitem[{Gitman and Ginsburg(2017)}]{Gitman-17}
Gitman, I., Ginsburg, B.: Comparison of batch normalization and weight
  normalization algorithms for the large-scale image classification.
\newblock CoRR abs/1709.08145 (2017)

\bibitem[{Graves(2011)}]{Graves-11}
Graves, A.: Practical variational inference for neural networks.
\newblock In: NIPS, pp. 2348--2356 (2011)

\bibitem[{He et~al.(2016)He, Zhang, Ren, and Sun}]{HeZRS16}
He, K., Zhang, X., Ren, S., Sun, J.: Deep residual learning for image
  recognition.
\newblock In: CVPR. pp. 770--778 (2016)

\bibitem[{Ioffe and Szegedy(2015)}]{IoffeS15}
Ioffe, S., Szegedy, C.: Batch normalization: Accelerating deep network training
  by reducing internal covariate shift.
\newblock In: ICML. vol.~37, pp. 448--456 (2015)

\bibitem[{Kingma et~al.(2015)Kingma, Salimans, and Welling}]{Kingma-15-dropout}
Kingma, D.P., Salimans, T., Welling, M.: Variational dropout and the local
  reparameterization trick.
\newblock In: NIPS, pp. 2575--2583 (2015)

\bibitem[{Lee(2012)}]{lee2012bayesian}
Lee, P.: Bayesian Statistics: An Introduction (2012)

\bibitem[{{Lei Ba} et~al.(2016){Lei Ba}, {Kiros}, and
  {Hinton}}]{Ba-2016-Layer-Norm}
{Lei Ba}, J., {Kiros}, J.R., {Hinton}, G.E.: {Layer Normalization}.
\newblock ArXiv e-prints  (Jul 2016)

\bibitem[{Li et~al.(2018)Li, Chen, Hu, and Yang}]{XiangLi-18-BN}
Li, X., Chen, S., Hu, X., Yang, J.: Understanding the disharmony between
  dropout and batch normalization by variance shift.
\newblock CoRR abs/1801.05134 (2018)

\bibitem[{Luenberger and Ye(2015)}]{Luenberger:2015:LNP}
Luenberger, D.G., Ye, Y.: Linear and Nonlinear Programming (2015)

\bibitem[{Maška and \etal(2014)}]{cell-tracking-algs}
Maška, M., \etal: A benchmark for comparison of cell tracking algorithms.
\newblock Bioinformatics 30(11), 1609--1617 (2014)

\bibitem[{Ronneberger et~al.(2015)Ronneberger, Fischer, and Brox}]{U-net}
Ronneberger, O., Fischer, P., Brox, T.: {U-Net}: Convolutional networks for
  biomedical image segmentation.
\newblock In: MICCAI. pp. 234--241 (2015)

\bibitem[{Salimans and Kingma(2016)}]{Salimans2016WeightNA}
Salimans, T., Kingma, D.P.: Weight normalization: A simple reparameterization
  to accelerate training of deep neural networks.
\newblock In: NIPS (2016)

\bibitem[{Santurkar et~al.(2018)Santurkar, Tsipras, Ilyas, and
  Madry}]{Santurkar-18}
Santurkar, S., Tsipras, D., Ilyas, A., Madry, A.: How does batch normalization
  help optimization? (no, it is not about internal covariate shift).
\newblock CoRR 1805.11604 (2018)

\bibitem[{Schulman et~al.(2015)Schulman, Heess, Weber, and
  Abbeel}]{Schulman-15}
Schulman, J., Heess, N., Weber, T., Abbeel, P.: Gradient estimation using
  stochastic computation graphs.
\newblock In: NIPS, pp. 3528--3536 (2015)

\bibitem[{Shekhovtsov and Flach(2018)}]{shekhovtsov-18-norm}
Shekhovtsov, A., Flach, B.: Normalization of neural networks using analytic
  variance propagation.
\newblock In: Computer Vision Winter Workshop. pp. 45--53 (2018)

\bibitem[{Springenberg et~al.(2015)Springenberg, Dosovitskiy, Brox, and
  Riedmiller}]{Springenberg-14}
Springenberg, J., Dosovitskiy, A., Brox, T., Riedmiller, M.: Striving for
  simplicity: The all convolutional net.
\newblock In: ICLR (workshop track) (2015)

\bibitem[{Srivastava et~al.(2014)Srivastava, Hinton, Krizhevsky, Sutskever, and
  Salakhutdinov}]{srivastava14a}
Srivastava, N., Hinton, G., Krizhevsky, A., Sutskever, I., Salakhutdinov, R.:
  Dropout: A simple way to prevent neural networks from overfitting.
\newblock JMLR 15, 1929--1958 (2014)

\bibitem[{Teye et~al.(2018)Teye, Azizpour, and Smith}]{Teye-18}
Teye, M., Azizpour, H., Smith, K.: Bayesian uncertainty estimation for batch
  normalized deep networks.
\newblock In: ICML (2018)

\bibitem[{Zagoruyko and Komodakis(2016)}]{Zagoruyko-16-WRN}
Zagoruyko, S., Komodakis, N.: Wide residual networks.
\newblock In: BMVC. pp. 87.1--87.12 (September 2016)

\end{thebibliography}
\newpage
\appendix
\numberwithin{figure}{section}
\addtocontents{toc}{\protect\setcounter{tocdepth}{2}}
\pagestyle{plain}
%
\setcounter{figure}{0}
\setcounter{table}{0}
\counterwithin{figure}{section}
\counterwithin{table}{section}
\counterwithin{theorem}{section}
\counterwithin{proposition}{section}
\counterwithin{lemma}{section}
%
\title{Appendix}
\author{}
\institute{}
\titlerunning{\,}
\authorrunning{\,}
\maketitle
%
%
%
%
\let\Contentsline\contentsline
\renewcommand\contentsline[3]{\Contentsline{#1}{#2}{}}
\makeatletter
\renewcommand{\@dotsep}{10000}
\makeatother
\def\authcount#1{}
%

\section{Additional Experiments}\label{sec:experiment}
\subsection{CIFAR}\label{sec:CIFAR}
CIFAR10\footnote{\url{https://www.cs.toronto.edu/~kriz/cifar.html}} is a common vision benchmark for new methods. Unlike the very popular MNIST dataset, its classification performance is not yet saturated.
From the training set we split 10 percent (at random) to create a validation set. The validation set is meant for model selection and monitoring the validation loss and accuracy during learning. The test set was kept for the final evaluation only.

The all CNN network we test~\cite{Springenberg-14} has the following structure of convolutional layers:
\begin{lstlisting}
ksize = [3,  3,  3,  3,   3,   3,   3,   1,   1 ]
stride= [1,  1,  2,  1,   1,   2,   1,   1,   1 ]
depth = [96, 96, 96, 192, 192, 192, 192, 192, 10]
\end{lstlisting}
each but the last one ending with leaky ReLU activation with leaky slope 0.01. The final layers of the network are
\begin{lstlisting}
Norm, AdaptiveAvgPool2d, LogSoftmax
\end{lstlisting}
When we train with either of the normalization variants, it is introduced after each convolutional layer.

For the optimization in all experiments we used batch size 32 (which was determined as optimal with a manual grid search), SGD optimizer with Nesterov Momentum 0.9 (pytorch default) and the learning rate $lr \cdot \gamma^{k}$, where $k$ is the epoch number, $lr$ is the initial learning rate, $\gamma$ is the decrease factor. In all reported results for CIFAR we used $\gamma$ such that $\gamma^{600}$ = 0.1 and 1200 epochs. This relatively longer training schedule is used in order to make comparison in terms of accuracy more fair across somewhat faster and somewhat slower methods. The initial learning rate was selected by an automatic numerical search optimizing the training loss achieved in 5 epochs. This is performed individually per training case to take care for the differences introduced by different reparametrizations. 

Parameters of linear and convolutional layers were initialized using pytorch defaults, \ie, uniformly distributed in $[-1/\sqrt{c},\, 1/\sqrt{c}]$, where c is the number of inputs per one output. Standard training and weight normalization were additionally initialized with data-dependent normalization equivalent to making one pass with batch normalization with zero learning rate, batch size 128 and applying the scaling and bias parameters computed by this pass. This is similar to~\cite{Gitman-17,shekhovtsov-18-norm} and ensures that these methods start from the same initialization in comparison plots with BN. Analytic normalization~\cite{shekhovtsov-18-norm} uses analytic approximate statistics for initialization.

Standard minor data augmentation was applied to the training and validation sets, consisting in random translations $\pm 2$ pixels (with zero padding) and horizontal flipping.

\subsubsection{Dependence on Batch Size}
\begin{figure}[!t]
\centering
\setlength{\tabcolsep}{2pt}
\resizebox{\linewidth}{!}{
\begin{tabular}{cc}
\ \ \ \ \footnotesize{Training Loss} & \ \ \ \ \footnotesize Validation Loss \\
\includegraphics[width=0.5\linewidth]{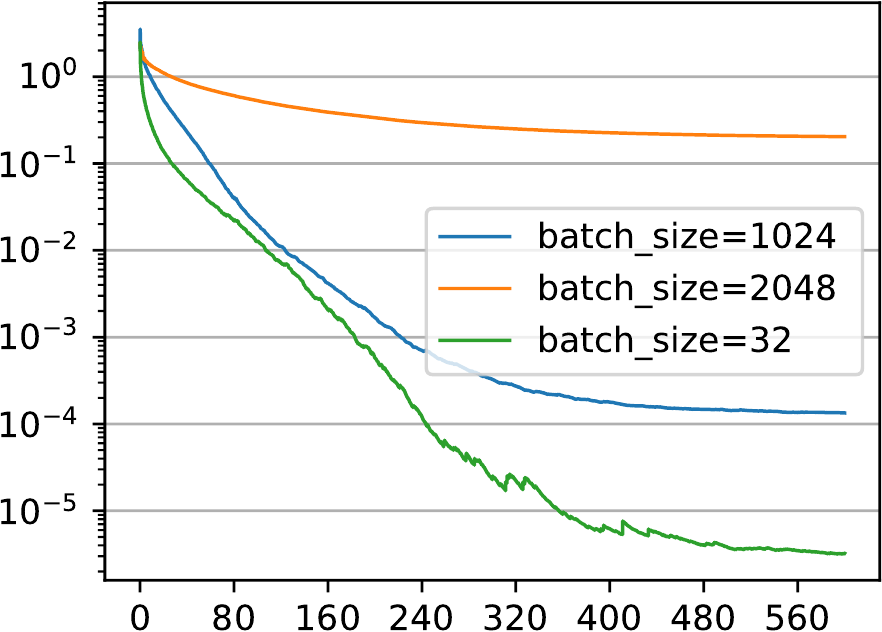}&%
\includegraphics[width=0.5\linewidth]{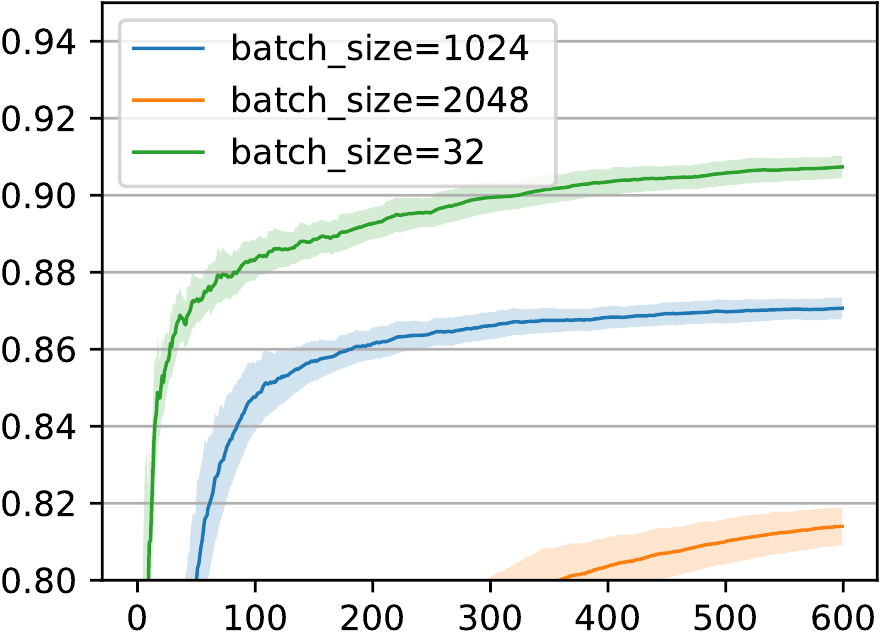}\\
\includegraphics[width=0.5\linewidth]{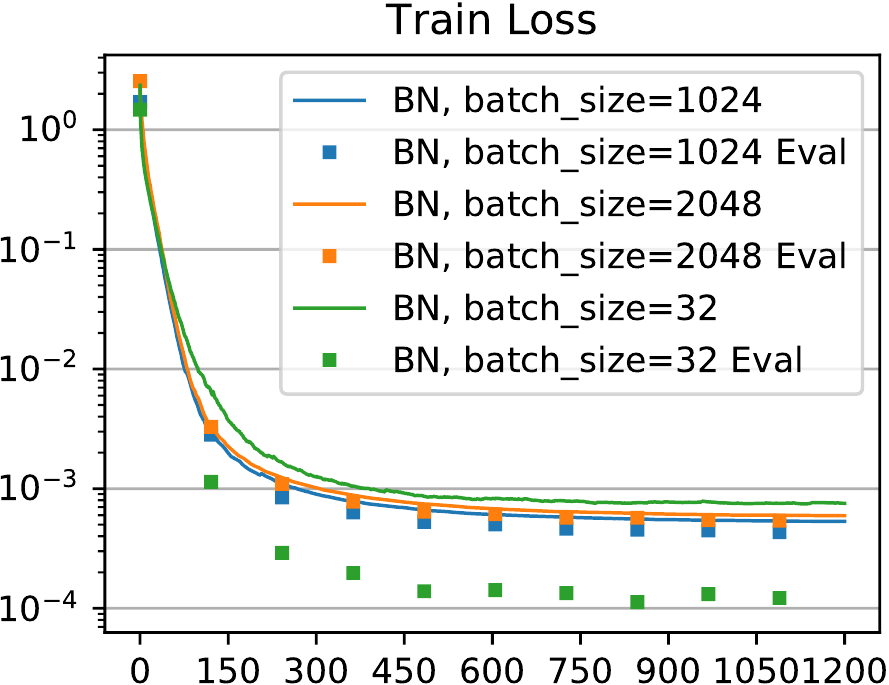}&%
\includegraphics[width=0.5\linewidth]{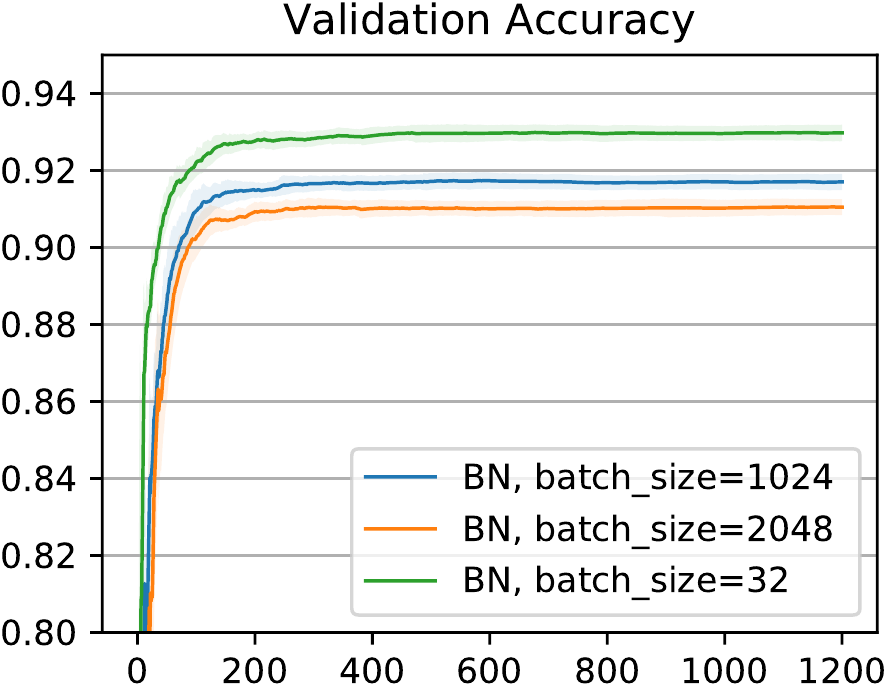}\\
\end{tabular}
}
\caption{Dependence of SGD on training batch size. {\em Top:} standard training. {\em Bottom:} batch normalization. All cases use the same initialization point, the learning rate is automatically found for each run. We observe that with or without BN, the results of SGD significantly depend on the batch size.
\label{fig:batch-dependance}}
\end{figure}

\cref{fig:batch-dependance} shows dependence of SGD on the batch size. It is widely observed that SGD has a faster convergence to the region of interest than the full gradient descent. Similarly, when increasing the batch size, we observe that the learning slows down. Moreover, stochasticity of the SGD and its regularization effect on the learning is changed. We observe this with BN normalization: although optimizing the training loss succeeds, the stochasticity of both BN and SGD are decreased leading to a significant loss of accuracy.
\subsubsection{Statistical Significance / Reproducibility}
\begin{figure}
\centering
\setlength{\tabcolsep}{2pt}
\resizebox{\linewidth}{!}{
\begin{tabular}{cc}
\ \ \ \ \footnotesize{Batch Normalization} & \ \ \ \ \footnotesize Analytic normalization with Bayesian learning\\
\includegraphics[width=0.5\linewidth]{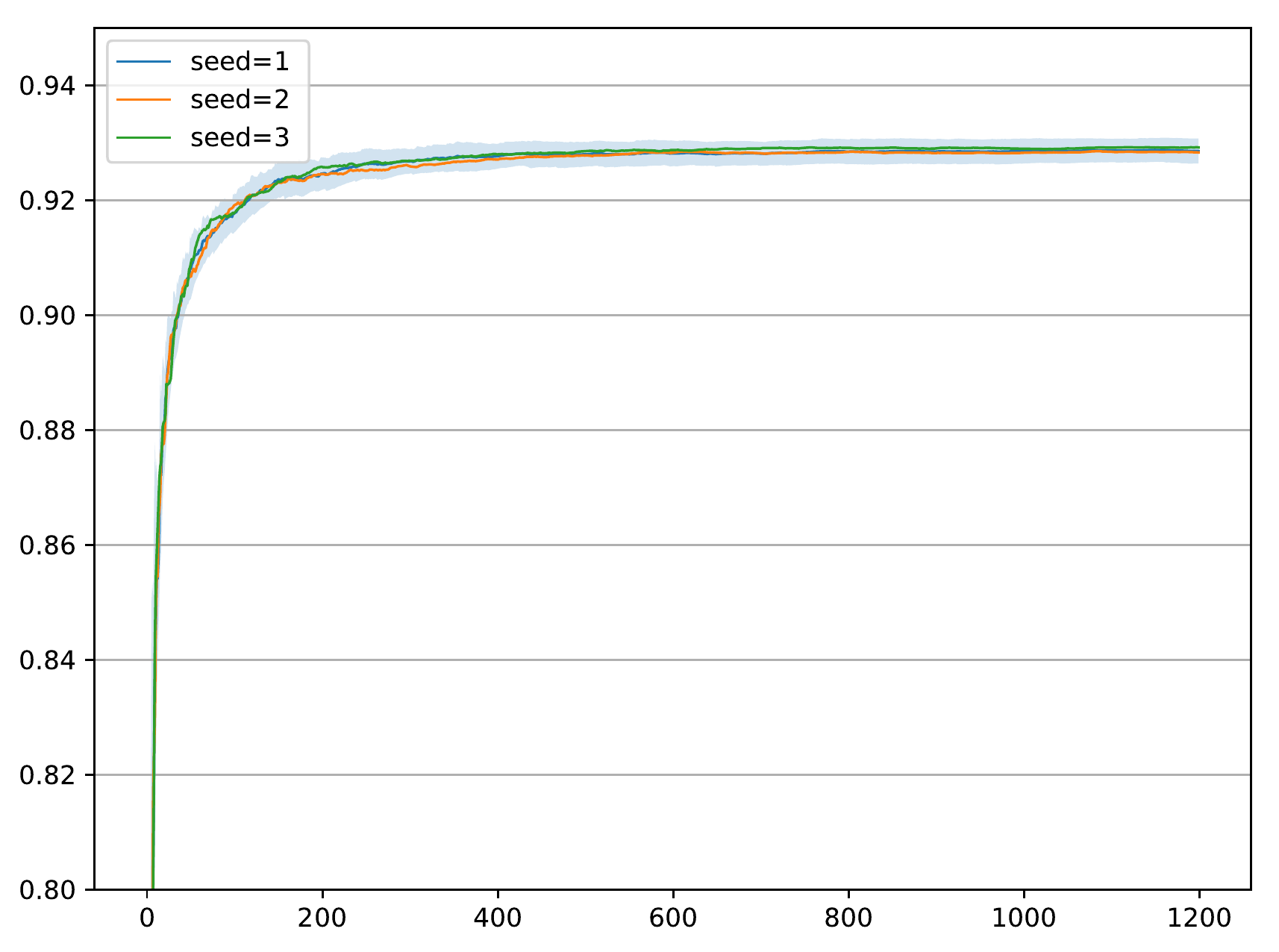}&%
\includegraphics[width=0.5\linewidth]{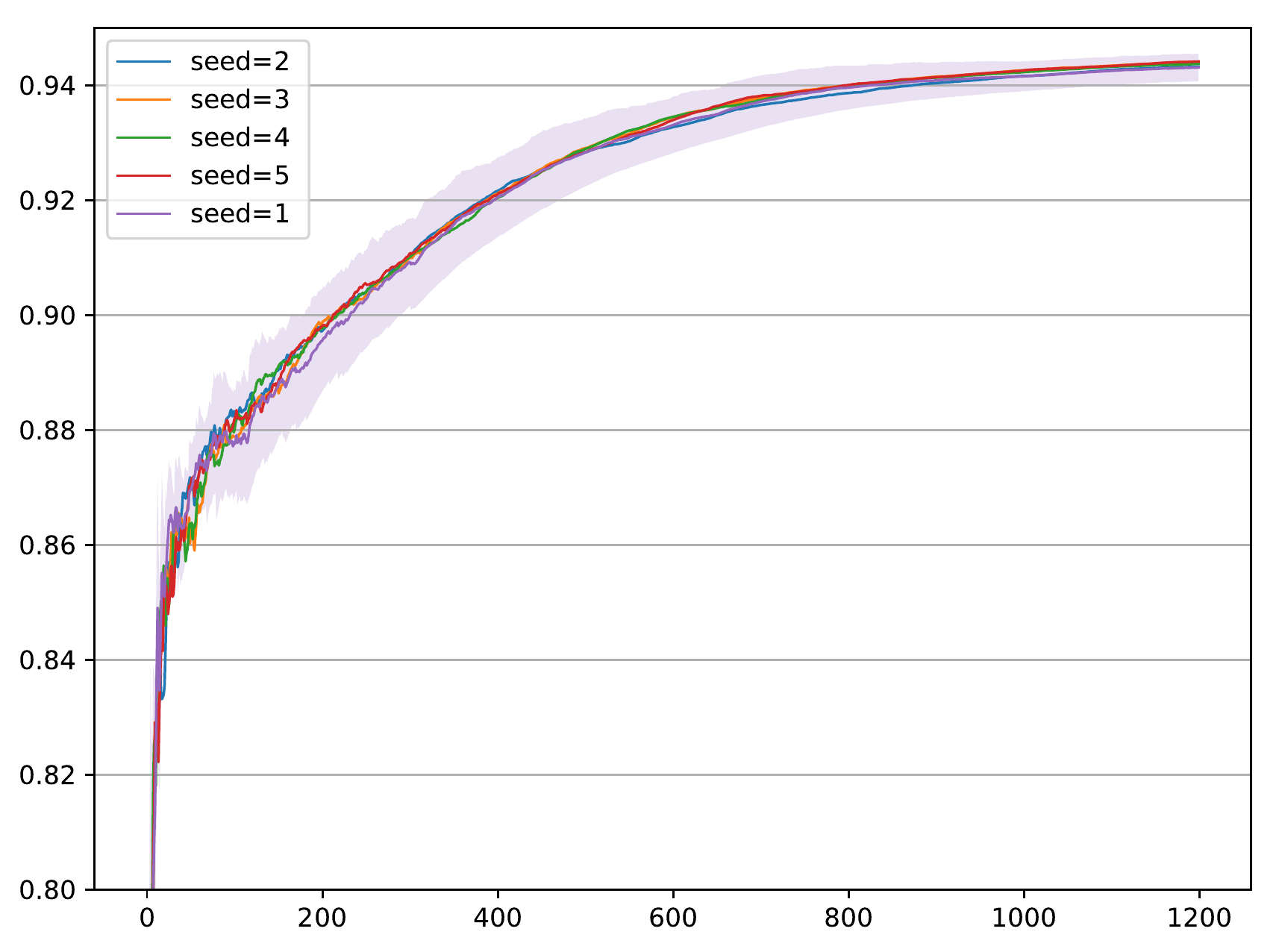}\\%
\end{tabular}
}
\caption{Validation accuracy vs. epochs for runs with different random seeds.
Solid lines show (running) mean of each run and shaded area shows standard deviation of iterates of the run with seed 1 (corresponding to the results shown in the main paper).
\label{fig:seeds}
}
\end{figure}

A fair question about the reported comparisons is their statistical significance. After all, the initialization, batch selection and parameter samples in Bayesian learning are all random. 
Despite of that, the results are fairly repeatable. \cref{fig:seeds} shows the evidence for BN and the proposed method: runs with different random seeds are within the standard deviation of iterates of a single run. In this experiment the split of the training data into training and validation sets and the learning rate are kept constant. We observe a similar behavior with other studied methods and metrics. We therefore show only the standard deviation of the iterates of a single run in all our validation plots in the main paper (as shaded areas).
\subsubsection{Recognition with a Reject Option}
\begin{figure}
\centering
\setlength{\tabcolsep}{2pt}
\begin{tabular}{cc}
\includegraphics[width=0.5\linewidth]{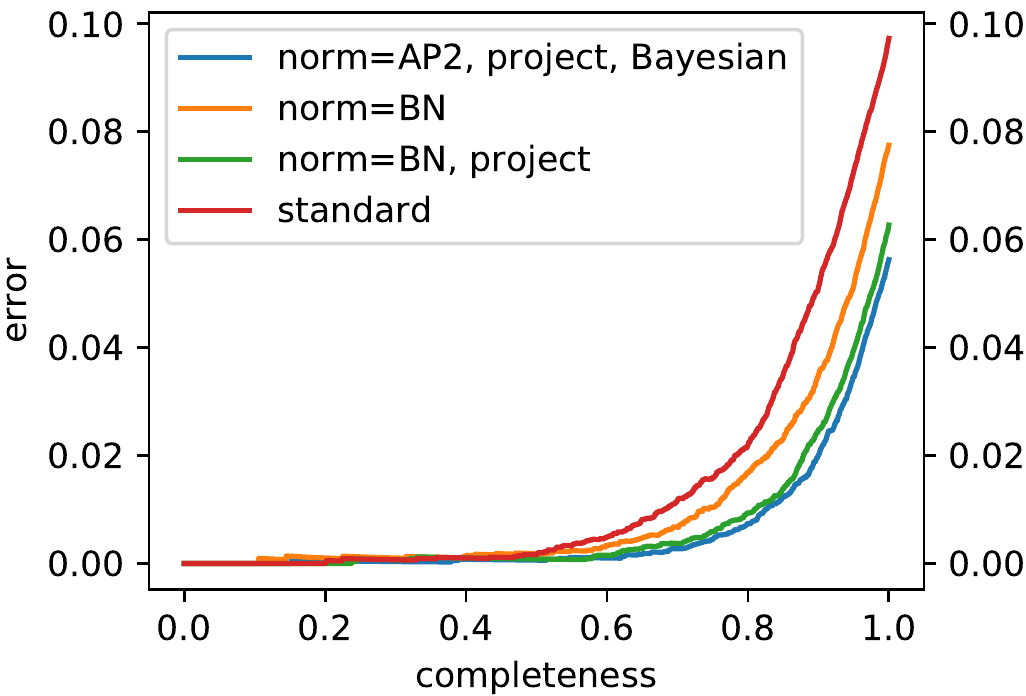}&%
\end{tabular}
\caption{Error coverage test. By rejecting from recognition on the data with high entropy of the predictive distribution, the recognition of the accepted data can be much more accurate. This verifies that the learned posterior distribution $p(y|x)$ contains useful confidence information -- the recognition system can know when the prediction is likely to be erroneous and when not.
\label{fig:coverage}
}
\end{figure}

Except of the bare accuracy reported in Table~\cref{tab:results-test} we conducted further tests to see the quality of the learned predictive distribution. \cref{fig:coverage} shows the {\em error coverage} plot obtained as follows. The recognition system is allowed to reject from recognition based on a certain criterion, \ie to give an ``I don't know" answer. The error rate of the classified data (error) is plotted versus the portion of classified data to all data (completeness). As a criterion for rejection, for simplicity, we use the entropy of the predictive distribution $p(y|x)$, \ie, samples $x$ with high entropy are first candidates for "I don't know" answer.
The results are obtained with deterministic ``single run" methods (\ie, without sampling batches or stochastic parameters) on the test set.
It is seen that regularization uniformly improves the recognition accuracy at all thresholds and that the proposed method is on par with batch normalization while potentially more flexible. It is interesting to note that close to zero error rates are possible with the completeness threshold of about 60\% of the cases.

\subsubsection{Sensitivity to Input Perturbations}
\begin{figure}
\centering
\setlength{\tabcolsep}{2pt}
\begin{tabular}{cc}
\includegraphics[width=0.5\linewidth]{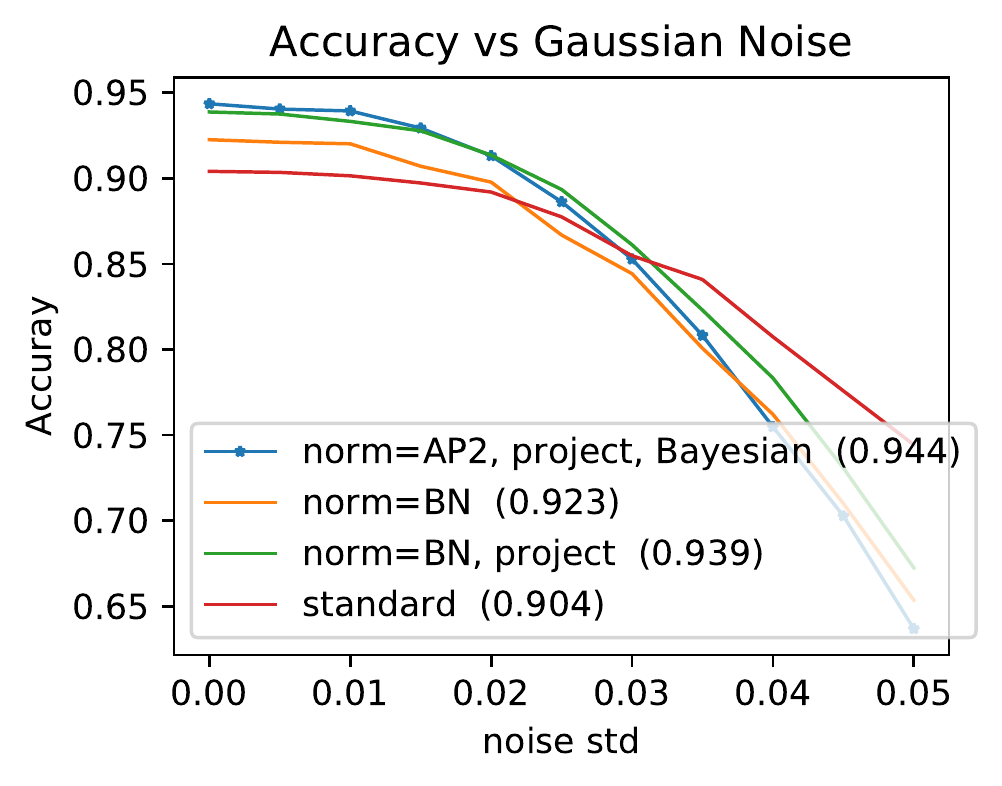}&%
\includegraphics[width=0.5\linewidth]{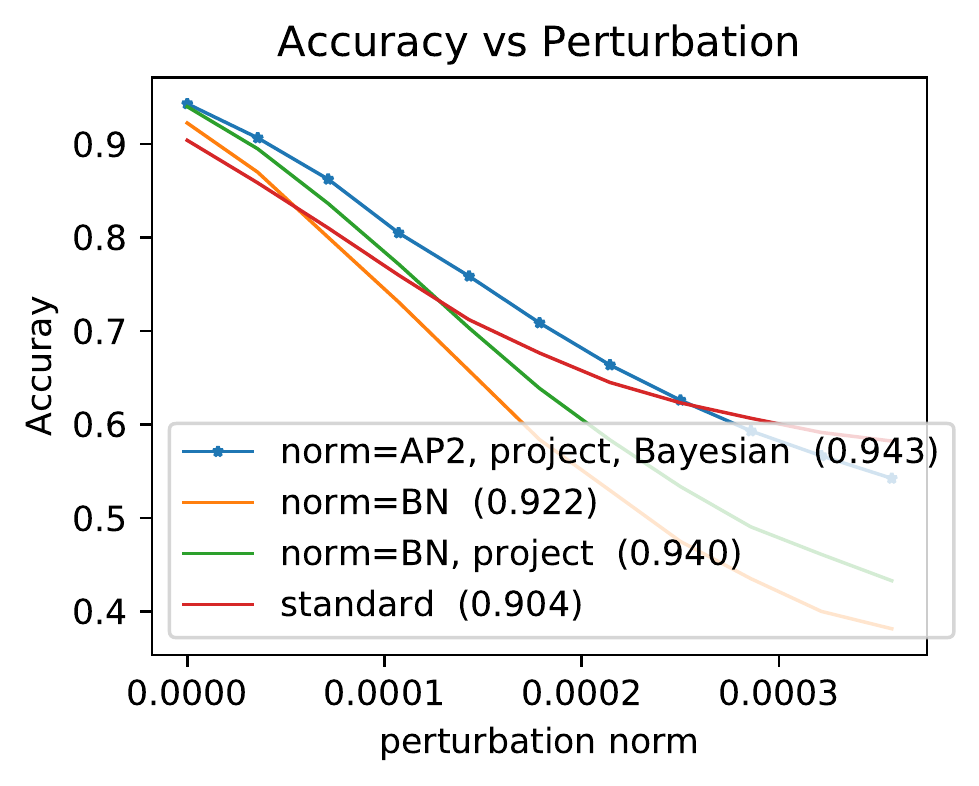}%
\end{tabular}
\caption{Stability with respect to perturbation of the input. {\em Left:} the input image is perturbed with a Gaussian noise with a certain standard deviation (x-axis).
{\em Right:} the input image is perturbed with an adversarial gradient sign attack of a certain norm of the perturbation (x-axis). Numbers in brackets show accuracy at zero noise.
\label{fig:stability}
}
\end{figure}

\cref{fig:stability} inspects the sensitivity of the learned models to the perturbation of the input with a random noise or with an adversarial gradient sign attack. The tests reveal relatively similar behavior of all solutions experiencing a fast drop of accuracy when the perturbation strength increases. Note that deterministic test-time models are evaluated, \ie, no sampling of batches or parameters at test time. The test shows that stochastic regularization in the form of Bayesian learning did not improve stability of such deterministic predictions.

\subsection{Cell Segmentation}\label{sec:hela-supp}

This experiment is conducted on the dataset "DIC-HeLa"~\cite{cell-tracking-algs} which contains 20 fully annotated training images of the size 512x512. We randomly split the training set into 70\% training and 30\% validation. We restrict ourselves to the task of segmenting the cells.

The evaluated network is a fully convolutional network with the following structure:
\begin{lstlisting}
Conv2D(channels=64, ksize=9), Activation
Conv2D(channels=64, ksize=9), Activation
AvgPool2d(ksize=3)
Conv2D(channels=32, ksize=9), Activation
Conv2D(channels=32, ksize=9), Activation
AvgPool2d(ksize=3)
Conv2D(channels=16, ksize=9), Activation
Conv2D(channels=16, ksize=9), Activation
AvgPool2d(ksize=3)
Conv2D(channels=8, ksize=9), Activation
Conv2D(channels=8, ksize=9), Activation
AvgPool2d(ksize=3)
Conv2D(channels=4, ksize=9), Activation
Conv2D(channels=4, ksize=9), Activation
AvgPool2d(ksize=3)
Conv2D(channels=2, ksize=1), Activation
LogSoftMax
\end{lstlisting}
As activation we used the Softmax function. This network has about $600k$ parameters, which is small compared to U-net~\cite{U-net} that has 30M parameters and has been previously applied to the small dataset.
Following~\cite{U-net}, we perform the following data augmentation: we consider random vertical and horizontal flips as well as small non-rigid deformations (the images and the corresponding segmentations are deformed synchronously).  The deformations are illustrated in~\cref{fig:hela-data}.

For learning of this large network, Adam optimizer was more suitable. We used the same learning rate schedule $lr \cdot \gamma^{k}$ and selection of initial learning rate as with CIFAR, but with $\gamma$ set such that $\gamma^{1000}$ = 0.1 and used 3000 epochs, which was appropriate for the amount of data augmentation that we applied.

\end{document}